\documentclass{article}

\usepackage{microtype}
\usepackage{xspace}
\usepackage{graphicx}
\usepackage{subcaption}
\usepackage{booktabs} 

\usepackage{hyperref}

\newcommand{\modelname}{MEL\xspace}

\usepackage[preprint]{icml2026}

\usepackage{amsmath}
\usepackage{amssymb}
\usepackage{mathtools}
\usepackage{amsthm}

\usepackage[capitalize,noabbrev]{cleveref}

\usepackage{relsize}
\usepackage{xcolor}
\usepackage[most]{tcolorbox}
\usepackage{listings}
\usepackage{paracol} 
\usepackage{amsmath}
\usepackage{algorithm}
\renewcommand{\algorithmicrequire}{\textbf{Input:}}
\renewcommand{\algorithmicensure}{\textbf{Output:}}
\usepackage{graphicx}
\usepackage{subcaption}	
\usepackage{multirow}
\usepackage{cuted}
\usepackage{marvosym}

\definecolor{bg}{RGB}{176,226,255}
\definecolor{bonus_green}{RGB}{0,100,0}

\usepackage{tabularx}
\usepackage[table,xcdraw,dvipsnames, svgnames, x11names]{xcolor}
\usepackage{booktabs}

\theoremstyle{plain}

\theoremstyle{definition}

\theoremstyle{remark}

\usepackage[textsize=tiny]{todonotes}

\icmltitlerunning{Preprint}

\begin{document}

\twocolumn[
  \icmltitle{Internalizing Meta-Experience into Memory for Guided \\
  Reinforcement Learning in Large Language Models}



  \icmlsetsymbol{corresponding}{\Letter}

  \begin{icmlauthorlist}
    \textbf{Shiting Huang}$^{1}$\,
    \textbf{Zecheng Li}$^{1}$\,
    \textbf{Yu Zeng}$^{1}$\,
    \textbf{Qingnan Ren}$^{1}$\,
    \textbf{Zhen Fang}$^{1}$\,
    \textbf{Qisheng Su}$^{1}$\, \\
    \textbf{Kou Shi}$^{1}$\,
    \textbf{Lin Chen}$^{1}$\,
    \textbf{Zehui Chen}$^{1}$\,
    \textbf{Feng Zhao}$^{1}$\textsuperscript{\Letter} \\
    $^1$University of Science and Technology of China \\
    \Letter: Corresponding Author
  \end{icmlauthorlist}
  
  \vskip 0.3in
  \printAffiliationsAndNotice{}
]

\begin{abstract} 
Reinforcement Learning with Verifiable Rewards (RLVR) has emerged as an effective approach for enhancing the reasoning capabilities of Large Language Models (LLMs).
Despite its efficacy, RLVR faces a meta-learning bottleneck: it lacks mechanisms for error attribution and experience internalization intrinsic to the human learning cycle beyond practice and verification, thereby limiting fine-grained credit assignment and reusable knowledge formation.
We term such reusable knowledge representations derived from past errors as meta-experience.
Based on this insight, we propose \textbf{M}eta-\textbf{E}xperience \textbf{L}earning (\textbf{\modelname}), a novel framework that incorporates self-distilled meta-experience into the model's parametric memory.
Building upon standard RLVR, we introduce an additional design that leverages the LLM's self-verification capability to conduct contrastive analysis on paired correct and incorrect trajectories, identify the precise bifurcation points where reasoning errors arise, and summarize them into generalizable meta-experience.
The meta-experience is further internalized into the LLM's parametric memory by minimizing the negative log-likelihood, which induces a language-modeled reward signal that bridges correct and incorrect reasoning trajectories and facilitates effective knowledge reuse.
Experimental results demonstrate that \modelname achieves consistent improvements on benchmarks, yielding 3.92\%--4.73\% Pass@1 gains across varying model sizes.
\end{abstract}
\section{Introduction}
Reinforcement Learning (RL) has emerged as a pivotal paradigm for enhancing the reasoning capabilities of Large Language Models (LLMs) on complex tasks, such as mathematics, programming, and logic reasoning~\cite{shao2024deepseekmath, chen2025enigmata, zeng2025agentic, wang2025vrag, zeng2025enhancing, zeng2026vision, huang2026vision}.
By leveraging feedback signals obtained from interaction with the task environment, RL enables LLMs to move beyond passive imitation learning toward goal-directed reasoning and action~\cite{schulman2017proximal,ouyang2022training,wulfmeier2024imitating}.
Furthermore, by replacing learned reward models with programmatically verifiable signals, Reinforcement Learning with Verifiable Rewards (RLVR) eliminates the need for expensive human annotations and mitigates reward hacking, thereby enabling models to explore problem-solving strategies more effectively, which has contributed to its growing attention~\cite{lambert2024tulu}.

However, RLVR still faces a fundamental bottleneck regarding the granularity and utilization of learning signals. 
From a meta-learning perspective, the human learning cycle involves three critical components: practice and verification, error attribution, and experience internalization. 
While RLVR primarily drives policy updates through practice and verification, it overlooks the critical stages of error attribution and experience internalization, both of which are essential for fine-grained credit assignment and the formation of reusable knowledge~\cite{wu2025evolver, zhang2025agent}.
In other words, RLVR is largely limited to assessing the overall quality of entire trajectories, while struggling to reason about fine-grained knowledge at the level of intermediate steps~\cite{xie2025capo}.
Although RL approaches~\cite{lightman2023let, khalifa2025process} employing Process Reward Models (PRMs) to provide dense learning signals attempt to mitigate this limitation, their reliance on trained proxies makes them inherently susceptible to reward hacking~\cite{cheng2025stop,guo2025deepseek}, and poses a fundamental tension with the RLVR paradigm, which is centered on programmatically verifiable rewards.

\begin{figure}[th!]
\centering
\includegraphics[width=1\columnwidth]{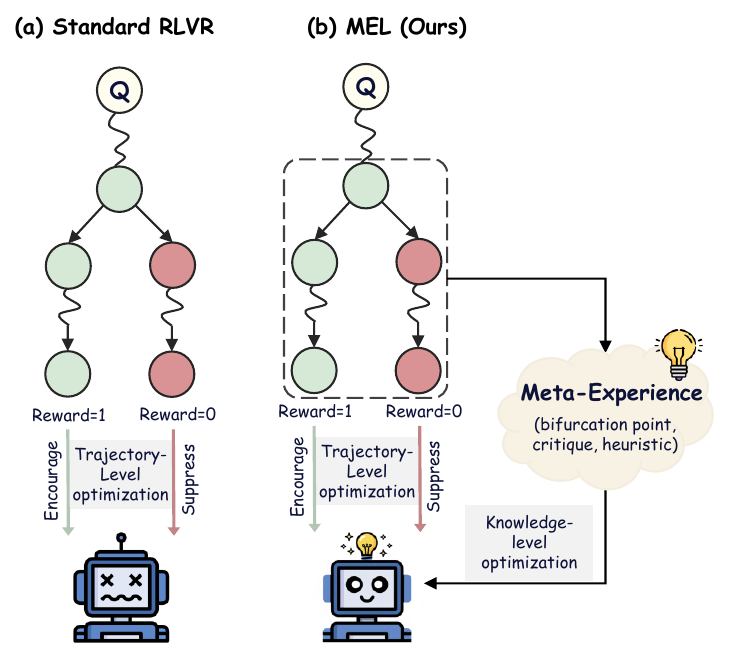}
  \caption{Paradigm comparison between standard RLVR and \modelname, where \modelname extends RLVR with an explicit knowledge-level learning loop.}
  \label{fig:intro}
\end{figure}

Recently, a growing number of studies have explored integrating experience learning within the RLVR framework to address the above challenge.
Early attempts, such as StepHint~\cite{zhang2025stephint} utilizes experience as hints to elicit superior reasoning paths from the original problems, treating these trajectories as off-policy migration signals.
However, the resulting off-policy deviation in response distribution can compromise optimization stability, undermining the theoretical benefits of on-policy reinforcement learning.
To alleviate such instability, Scaf-GRPO~\cite{zhang2025scaf} leverages superior models to generate multi-level knowledge-intensive experience, injecting them as on-policy prefixes for policy updates. 
Yet, while effective in teaching models to reason within specific experience-augmented distributions, such prefixes are unavailable during inference, inducing a severe distributional mismatch, thereby limiting performance gains.
Critically, despite their differences, these approaches utilize retrieved experience primarily as external hints.
While these strategies effectively elicit better reasoning paths during training, the resulting learning signals remain predominantly at the \textbf{\emph{trajectory-level}}, yielding superficial corrections rather than intrinsic cognitive enhancements.

Building on this insight, we introduce the concept of \emph{meta-experience}, elevating experience learning from trajectory-level instances to knowledge-level representations.
Through contrastive analysis on paired correct and incorrect trajectories, we pinpoint the bifurcation points underlying reasoning failures and abstracts them into reusable meta-experiences.
Accordingly, we propose \textbf{M}eta-\textbf{E}xperience \textbf{L}earning (\textbf{\modelname}), a framework explicitly designed to enable knowledge-level internalization and reuse of meta-experiences.
During training phase, \modelname leverages meta-experiences to inject generalizable insights via a self-distillation mechanism, and internalizes them by minimizing the negative log-likelihood in the model's parametric memory.
As shown in Figure~\ref{fig:intro}, \modelname differs from standard RLVR, which relies on coarse-grained outcome rewards and treats correct and incorrect trajectories independently, by explicitly connecting them via meta-experiences.
Hence, this process can be viewed as a language-modeled process-level reward signal, providing continuous and fine-grained guidance for improving reasoning capability.
To further enhance stability and effectiveness during RLVR training, we propose empirical validation via replay, which uses meta-experiences as auxiliary in-context hints to assess their contribution to output accuracy.
Meta-experiences that pass validation are integrated via negative log-likelihood minimization, while those that fail validation are excluded.
In summary, our main contributions are as follows:

\begin{itemize}

\item We propose \modelname, a novel framework that integrates self-distilled meta-experience with reinforcement learning, addressing the limitations of standard RLVR in error attribution and experience internalization by embedding these meta-experiences directly into the parametric memory of LLMs.

\item We validate the effectiveness of \modelname through extensive experiments on five challenging mathematical reasoning benchmarks across multiple LLM scales (4B, 8B, and 14B). Compared with both the vanilla GRPO baseline and the corresponding base models, \modelname consistently improves performance across Pass@1, Avg@8, and Pass@8 metrics.

\item Empirical results confirm that \modelname seamlessly integrates with diverse paradigms (e.g., RFT, GRPO, REINFORCE++) to reshape reasoning patterns and elevate performance ceilings. Notably, these benefits exhibit strong scalability, becoming increasingly pronounced as model size expands.

\end{itemize}

\begin{figure*}[t]
  \centering
    \includegraphics[width=\textwidth]{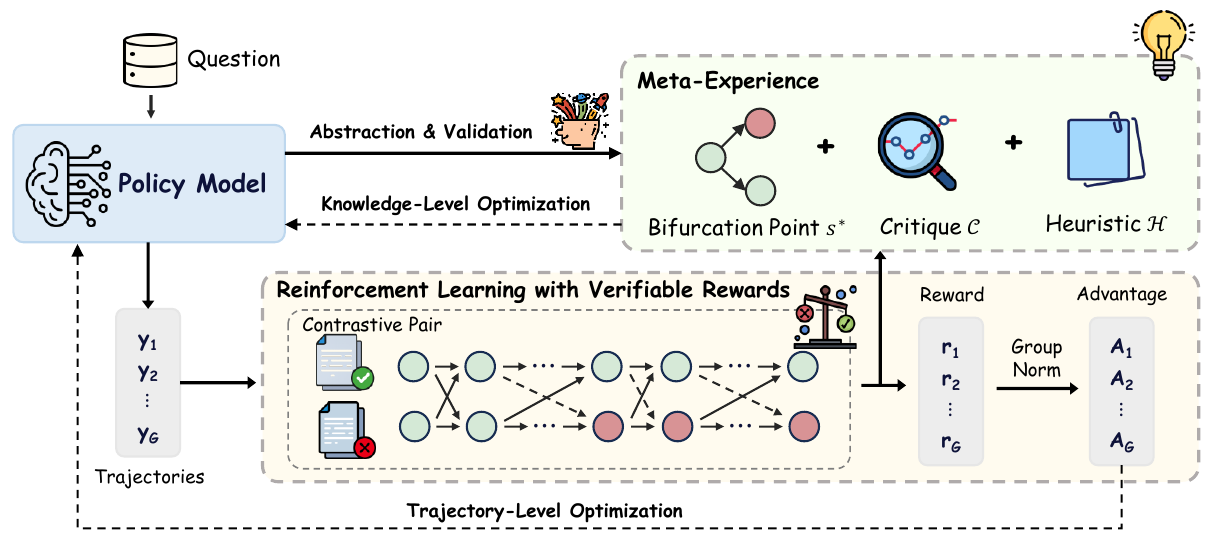}
  \caption{Overview of \textbf{M}eta-\textbf{E}xperience \textbf{L}earning (\modelname), which constructs meta-experiences from contrastive pairs via abstraction and validation, thereby introducing an explicit knowledge-level learning loop on top of standard RLVR.}
  \label{fig:framework}
\end{figure*}

\section{Related Work}
\paragraph{Reinforcement Learning with Verifiable Rewards.} 
Reinforcement Learning with Verifiable Rewards (RLVR) leverages rule-based validators to provide deterministic feedback on models' self-generated solutions~\cite{lambert2024tulu}.
Extensive research has systematically investigated RLVR, exploring how this paradigm improves the performance of complex reasoning~\cite{jaech2024openai,guo2025deepseek,liu2025reinforcement,zhang2025survey}. 
The pioneering framework Group Relative Policy Optimization (GRPO)~\cite{shao2024deepseekmath} estimates advantages via group-wise relative comparisons, eliminating the need for a separate value model.
Building on this base method, recent studies have introduced a range of algorithmic variants to improve training stability and efficiency. 
For instance, REINFORCE++~\citep{hu2025reinforce} enhances stability through global advantage normalization; DAPO~\cite{yu2025dapo} mitigates entropy collapse and improves reward utilization via relaxed clipping and dynamic sampling; and GSPO~\cite{zheng2025group} reduces gradient estimation variance with sequence-level clipping.
Despite these algorithmic advancements, a fundamental limitation persists: current RLVR methods predominantly rely on outcome-level rewards. 
This failure to assign fine-grained credit to specific knowledge points prevents the construction of reusable knowledge formation, fundamentally constraining the development of systematic and generalizable reasoning capabilities.

\paragraph{Experience Learning.} 
Recent studies have increasingly recognized that leveraging various forms of experience can substantially enhance LLM reasoning capabilities.
One prominent line of research lies in test-time scaling methods, which store experience in external memory pools.
For example, SpeedupLLM~\cite{pan2025can} appends memories of previously reasoning traces as experience to accelerate inference, while Training Free GRPO~\cite{cai2025training} and ReasoningBank~\cite{ouyang2025reasoningbank} distill accumulated experience into structured memory entries for retrieval-based augmentation.
However, these approaches rely on ever-growing external memory, preventing the experience from being truly internalized and thus failing to substantively enhance intrinsic reasoning capabilities.
Complementarily, another stream of research integrates experience directly into RL training as guiding signals.
Methods such as Scaf-GRPO \cite{zhang2025scaf} and StepHint \cite{zhang2025stephint} employ external models to generate experiential hints, which are injected as prefixes or migration signals, to guide the policy toward higher-quality trajectories.
Similarly, approaches like LUFFY \cite{yan2025learning} and SRFT \cite{fu2025srft} incorporate expert solution traces as additional experience.
Despite improving exploration efficiency, these methods primarily induce trajectory-level imitation.
Consequently, models become proficient at following specific patterns but fail to develop the meta-cognitive understanding required for establishing reusable knowledge structures.


\section{Meta-Experience Learning}
Human learning follows a recurrent cognitive cycle consisting of practice and verification, error attribution, and experience internalization, which in turn informs subsequent practice.
Motivated by this cognitive process, we define {meta-experience} for LLMs as generalizable and reusable knowledge derived from accumulated reasoning trials, capturing both underlying knowledge concepts and common failure modes.
Building on this notion, we propose \textbf{M}eta-\textbf{E}xperience \textbf{L}earning \textbf{(\modelname)}, a framework operating within the RLVR paradigm and expressly designed to internalize such self-distilled, knowledge-level insights into the model's parametric memory.
As illustrated in Figure~\ref{fig:framework}, we first formalize the model exploration stage in RLVR (\S\ref{sec:rollout}), then present the details of the Meta-Experience construction (\S\ref{sec:construction}). Finally, we describe the internalization mechanism (\S\ref{sec:internalization}) for consolidating these insights into parametric memory, followed by the joint training objective for policy optimization (\S\ref{sec:objective}).

\subsection{Explorative Rollout and Verifiable Feedback}
\label{sec:rollout}
Mirroring the ``practice and check" phase in human learning, the RLVR framework engages the model in exploring potential solutions for reasoning tasks, while the environment serves as a deterministic verifier that provides verifiable feedback on the final answers.
As mastering complex logic typically requires traversing the solution space through multiple attempts, we simulate this stochastic exploration by adopting the group rollout formulation from Group Relative Policy Optimization (GRPO)~\cite{shao2024deepseekmath}.

Formally, given a query $x$ sampled from the dataset $\mathcal{D}$, the policy model $\pi_{\theta}$ performs stochastic exploration over the solution space and generates a group of $G$ independent reasoning trajectories
\(
\mathcal{Y} = \{y_1, y_2, \dots, y_G\}.
\)
A rule-based verifier then evaluates each trajectory using a verification function $V(\cdot)$, which compares the extracted final answer from $y_i$ against the ground-truth answer $y^*$ and assigns a binary outcome reward:
\begin{equation}
    r_i = \mathbb{I}\big[V(y_i, y^*)\big] \in \{0, 1\}.
\end{equation}

This process partitions the generated group $\mathcal{Y}$ into two distinct subsets: the set of correct trajectories $\mathcal{Y}^+ = \{y_i \mid r_i = 1\}$ and the set of incorrect trajectories $\mathcal{Y}^- = \{y_i \mid r_i = 0\}$.

The coexistence of $\mathcal{Y}^+$ and $\mathcal{Y}^-$ under the same prompt distribution suggests that the model is capable of solving the task, while producing diverse reasoning trajectories. 
For our method, such diversity constitutes a beneficial property and serves a dual role. 
On the one hand, it supplies the variance necessary for effective policy updates in standard RLVR. 
On the other hand, it enables the extraction of knowledge-level meta-expression through systematic contrast between correct and incorrect reasoning outcomes.

\subsection{Meta-Experience Construction}
\label{sec:construction}
Prior studies~\cite{xie2025capo, khalifa2025process, huang2025critictool} have shown that effective learning does not arise from merely knowing that a final answer is incorrect, but rather from identifying the specific bifurcation point at which the reasoning process deviates from the correct trajectory, a critical cognitive process known as error attribution.
Building on this insight, we leverage pairs of correct and incorrect trajectories to localize reasoning errors and distill such bifurcation points into explicit meta-experiences.

\paragraph{Locating the Bifurcation Point.}
To extract knowledge-level learning signals from the exploration results, we focus on identifying the bifurcation points where the reasoning logic diverges into an erroneous path.
With the exploration results partitioned into $\mathcal{Y}^+$ and $\mathcal{Y}^-$ by the verifier, we construct a set of contrastive pairs $\mathcal{P}_x = \{ (y^+, y^-) \mid y^+ \in \mathcal{Y}^+,\, y^- \in \mathcal{Y}^- \}$ for each query $x$, whose contrast naturally exposes the specific errors in the reasoning process.
Such contrastive analysis requires the presence of both positive and negative trajectories; accordingly, we only consider gradient-informative samples with non-empty $\mathcal{Y}^+$ and $\mathcal{Y}^-$.

For fine-grained comparison within each pair, each trajectory $y$ can be formatted as a reasoning chain $y = (s_1, s_2, \dots, s_L, a)$, where each $s_t$ denotes an atomic reasoning step and $a$ indicates the final answer.
Since both trajectories originate from the same context, they typically share a correct reasoning path until a critical divergence step $s^*$ occurs.

Given deterministic verification signals and full access to the reasoning chains, identifying the bifurcation point can be viewed as a discriminative task that is easier than reasoning from scratch~\cite{saunders2022self, swamy2025all}.
Motivated by this observation, we task the policy model with analyzing each contrastive pair to identify the reasoning bifurcation point $s^*$:
\begin{align}
    s^* &\sim \pi_{\theta}\big( \cdot \mid I, x, y^+, y^- \big).
\end{align}
Where $I$ denotes a structured instruction guiding introspective analysis.

\paragraph{Deep Diagnosis and Abstraction.}
Identifying the bifurcation point $s^*$ localizes where the reasoning fails, serving as the raw material for subsequent learning.
Anchored at $s^*$, the policy model conducts a deep diagnostic analysis to contrast the strategic choices underlying the successful and failed trajectories.
Specifically, the model examines the local reasoning context around $s^*$ to pinpoint the root cause of failure, such as violated assumptions, erroneous sub-goals, overlooked constraints, or the misuse of specific principles.
Complementarily, it inspects the successful trajectory to uncover the mechanisms that prevented such pitfalls, including precise knowledge application, explicit constraint verification, coherent knowledge representations, or emergent self-correction behaviors.
By jointly synthesizing these perspectives, the model distills the structural divergence between the correct and incorrect logic, crystallizing it into explicit knowledge.
Formally, we model this diagnostic process as generating a critique $\mathcal{C}$ that encapsulates the error attribution, the comparative strategic gap, and the corresponding corrective principle:
\begin{align}
    \mathcal{C} \sim \pi_{\theta}\big( \cdot \mid I, x, y^+, y^-, s^* \big).
\end{align}

To ensure generalization, it is imperative for the model to distill instance-specific critiques into abstract heuristics capable of guiding future reasoning.
This abstraction mechanism systematically strips away context-dependent variables, mapping the concrete logic of success and failure onto a generalized space of preconditions and responses.
Structurally, such heuristics synthesize abstract problem categorization with the corresponding reasoning principles, encompassing the essential knowledge points, theoretical theorems, and decision criteria.
Crucially, they also demarcate error-prone boundaries, explicitly highlighting potential pitfalls or latent constraints associated with the specific problem class.
We define the extraction of this heuristic knowledge $\mathcal{H}$ as a generation process conditioned on the full critique context:
\begin{equation}
    \mathcal{H} \sim \pi_{\theta}\big( \cdot \mid I, x, y^+, y^-, s^*, \mathcal{C} \big).
\end{equation}

Finally, we consolidate these components into a unified \emph{Meta-Experience} tuple $\mathcal{M}$, which elevates experience learning from trajectory-level instances to knowledge-level representations. 
\begin{equation}
    \mathcal{M} = \big( s^*, \mathcal{C}, \mathcal{H} \big).
\end{equation}

This formulation enables meta-experiences to be reused across tasks that share analogous reasoning structures, serving as a fine-grained learning signal.
By applying the extraction process across distinct contrastive pairs for a query $x$, we construct a candidate pool of meta-experiences $\mathcal{D}_{\mathcal{M}} = \{ (x, y^+_i, y^-_i, \mathcal{M}_i) \}_{i=1}^{N}$, where $N$ denotes the total number of meta-experiences derived from $x$, and $(y^+_i, y^-_i)$ represents the specific contrastive pair used to derive $\mathcal{M}_i$.

\paragraph{Empirical Validation via Replay.}
Closing the cognitive loop requires re-instantiating theoretical insights derived from past failures in future problem-solving to assess their validity.
We recognize that the raw meta-experience $\mathcal{M}$ may still suffer from intrinsic hallucinations or causal misalignment.
To mitigate this, we conduct empirical verification by incorporating the extracted tuple $\mathcal{M}$ as short-term working memory into the prompt, thereby guiding the model to re-attempt the original query $x$.
This procedure tests whether the injected meta-experience can effectively steer the model away from the previously identified bifurcation point $s^*$ and toward a correct reasoning trajectory.

We retain a meta-experience only if the corresponding replay trajectory $y_{\text{val}} \sim \pi_{\theta}(\cdot \mid x, \mathcal{M})$ satisfies the verifier by producing the correct answer.
\begin{equation}
\small
    \mathcal{D}_{\mathcal{M^*}} =
    \left\{
        (x, y^+, y^-, \mathcal{M}) \in \mathcal{D}_{\mathcal{M}}
        \;\middle|\;
        \mathbb{I}\big[V\!\left(y_{\text{val}}, y^*\right) = 1\big]
    \right\}.
\end{equation}

Consequently, this empirical validation preserves only high-quality meta-experiences for integration into parametric long-term memory, guaranteeing the reliability of the supervision signals used in the subsequent optimization phase.

\subsection{Internalization Mechanism}
\label{sec:internalization}

The verified meta-experiences $\mathcal{D}_{\mathcal{M}^*}$ constitute a high-quality reservoir of reasoning guidance.
However, treating these insights solely as retrieval-augmented memory imposes a substantial computational burden during the inference forward pass, as it necessitates processing elongated contexts for every query.
To overcome this limitation, we propose to transfer these insights from the transient context window to the model's parametric memory.
Unlike the finite context buffer, the model parameters offer a virtually unlimited capacity for accumulating diverse meta-experiences, allowing the policy to internalize vast amounts of reasoning patterns without incurring inference-time latency.

We establish this internalization process as a \emph{self-distilled paradigm}, where the model learns from its own verified experiences.
Specifically, we employ fine-tuning based on the token-averaged negative log-likelihood (NLL) objective to compile the meta-experiences into the policy's weights.
Formally, given the retrospective context $C_{\text{retro}}=[I, x, y^+, y^-]$, the internalization loss is defined as:
\begin{equation}
\small
\begin{aligned}
    \mathcal{L}_{\text{NLL}}(\theta) = - &\mathbb{E}_{(x, y^+, y^-, \mathcal{M}^*) \sim \mathcal{D}_{\mathcal{M}^*}} \\
    &\Big[\frac{1}{|\mathcal{M}^*|}\sum_{t=1}^{|\mathcal{M}^*|} \log \pi_{\theta}(\mathcal{M}^*_t \mid C_{\text{retro}}, \mathcal{M}^*_{<t}) \Big] \\
    = -& \mathbb{E}_{x\sim\mathcal{D},\,\{y_i\}_{i=1}^{G}\sim\pi_{\theta_{\mathrm{old}}}(\cdot\mid x)} \\
    & \Bigg[ \mathbb{E}_{(y^+,y^-,\mathcal{M}^*)\sim\mathcal{T}(x,\{y_i\}_{i=1}^{G})} \\
    & \Big[\frac{1}{|\mathcal{M}^*|}\sum_{t=1}^{|\mathcal{M}^*|} \log \pi_{\theta}(\mathcal{M}^*_t \mid C_{\text{retro}}, \mathcal{M}^*_{<t}) \Big] \Bigg]
\label{eq:nll}
\end{aligned}
\end{equation}
where $\mathcal{T}(\cdot)$ represents the stochastic construction process detailed in \S\ref{sec:construction}.

Based on this formulation, the internalization process can be viewed as a specialized sampling form within the RLVR framework.
By inverting the loss, we define the Meta-Experience Return $\mathcal{R}_{\text{MEL}}$ as the expected log-likelihood over the stochastically constructed verification set:
\begin{equation}
\small
\begin{aligned}
    \mathcal{R}_{\text{MEL}}& = \mathbb{E}_{(y^+,y^-,\mathcal{M}^*)\sim\mathcal{T}(x,\{y_i\}_{i=1}^{G})} \\
    &\Bigg[ \frac{1}{|\mathcal{M}^*|} \sum_{t=1}^{|\mathcal{M}^*|} \log \pi_{\theta}(\mathcal{M}^*_t \mid C_{\text{retro}}, \mathcal{M}^*_{<t}) \Bigg].
\label{eq:return}
\end{aligned}
\end{equation}

\subsection{Joint Training Objective}
\label{sec:objective}

\begin{table*}[t]
    \centering
    \small 
    \caption{\textbf{Main Results Comparison.} Comparison of Pass@1, Avg@8, and Pass@8 accuracy (\%) across different model scales. The best results within each model scale are marked in \textbf{bold}.}
    \label{tab:main_results_avg8}
    
    \resizebox{\textwidth}{!}{%
        \begin{tabular}{lccccccccc}
            \toprule
             & \multicolumn{3}{c}{\textbf{AIME 2024}} & \multicolumn{3}{c}{\textbf{AIME 2025}} & \multicolumn{3}{c}{\textbf{AMC 2023}} \\
            \cmidrule(lr){2-4} \cmidrule(lr){5-7} \cmidrule(lr){8-10}
            \textbf{Method} & \textbf{Pass@1} & \textbf{Avg@8} & \textbf{Pass@8} & \textbf{Pass@1} & \textbf{Avg@8} & \textbf{Pass@8} & \textbf{Pass@1} & \textbf{Avg@8} & \textbf{Pass@8} \\
            \midrule
            
            \multicolumn{10}{l}{\textit{\textbf{Qwen3-4B-Base}}} \\
            Baseline
            & 13.33 & 9.90 & 30.00 
            & 10.00 & 6.56 & 23.33 
            & 45.00 & 42.73 & 72.50 \\
            GRPO 
            & 13.33 & 18.33 & 30.00 
            & 6.67 & 17.50 & 30.00 
            & 57.50 & 58.13 & 85.00 \\
            \rowcolor{MintCream}
            \modelname
            & \textbf{20.00} & \textbf{20.83} & \textbf{33.00} 
            & \textbf{16.67} & \textbf{18.33} & \textbf{33.00} 
            & \textbf{60.00} & \textbf{60.31} & \textbf{87.50} \\
            
            \midrule
            
            \multicolumn{10}{l}{\textit{\textbf{Qwen3-8B-Base}}} \\
            Baseline
            & 6.67 & 10.00 & 26.67 
            & 13.33 & 15.00 & 33.33 
            & 65.00 & 52.50 & 87.50 \\
            GRPO 
            & 16.67 & 24.58 & 43.33 
            & 20.00 & 20.83 & \textbf{36.67} 
            & 67.50 & 69.06 & 87.50 \\
            \rowcolor{MintCream}
            \modelname
            & \textbf{30.00} & \textbf{25.42} & \textbf{60.00} 
            & \textbf{23.33} & \textbf{23.33} & \textbf{36.67} 
            & \textbf{70.00} & \textbf{70.31} & \textbf{90.00} \\

            \midrule
            
            \multicolumn{10}{l}{\textit{\textbf{Qwen3-14B-Base}}} \\
            Baseline
            & 13.33 & 10.83 & 36.67 
            & 6.66 & 9.58 & 33.33 
            & 60.00 & 51.25 & 82.50 \\
            GRPO 
            & 30.00 & 35.41 & 56.67 
            & 33.33 & 24.17 & 43.33 
            & 75.00 & 75.94 & \textbf{95.00} \\
            \rowcolor{MintCream}
            \modelname
            & \textbf{33.33} & \textbf{35.83} & \textbf{60.00} 
            & \textbf{36.67} & \textbf{30.00} & \textbf{46.67} 
            & \textbf{82.50} & \textbf{82.81} & \textbf{95.00} \\
            \bottomrule
        \end{tabular}
    }

    \vspace{5pt} 

    \resizebox{\textwidth}{!}{%
        \begin{tabular}{lccccccccc}
             & \multicolumn{3}{c}{\textbf{MATH 500}} & \multicolumn{3}{c}{\textbf{OlympiadBench}} & \multicolumn{3}{c}{\textbf{Average}} \\
            \cmidrule(lr){2-4} \cmidrule(lr){5-7} \cmidrule(lr){8-10}
            \textbf{Method} & \textbf{Pass@1} & \textbf{Avg@8} & \textbf{Pass@8} & \textbf{Pass@1} & \textbf{Avg@8} & \textbf{Pass@8} & \textbf{Pass@1} & \textbf{Avg@8} & \textbf{Pass@8} \\
            \midrule
            
            \multicolumn{10}{l}{\textit{\textbf{Qwen3-4B-Base}}} \\
            Baseline
            & 74.20 & 65.74 & 89.60 
            & 39.17 & 35.37 & 60.38 
            & 36.34 & 32.06 & 55.16 \\
            GRPO 
            & 81.80 & 82.20 & 93.00 
            & \textbf{48.51} & 48.46 & 67.21 
            & 41.56 & 44.92 & 61.04 \\
            \rowcolor{MintCream}
            \modelname
            & \textbf{82.20} & \textbf{82.30} & \textbf{93.80} 
            & \textbf{48.51} & \textbf{49.48} & \textbf{69.73} 
            & \textbf{45.48} & \textbf{46.25} & \textbf{63.41} \\
            
            \midrule
            
            \multicolumn{10}{l}{\textit{\textbf{Qwen3-8B-Base}}} \\
            Baseline
            & 77.00 & 73.40 & 91.40 
            & 44.51 & 39.41 & 64.09 
            & 41.30 & 38.06 & 60.60 \\
            GRPO 
            & 84.40 & 86.28 & 95.40 
            & 53.56 & 54.60 & \textbf{73.74} 
            & 48.43 & 51.07 & 67.33 \\
            \rowcolor{MintCream}
            \modelname
            & \textbf{86.60} & \textbf{86.70} & \textbf{96.20} 
            & \textbf{54.01} & \textbf{55.60} & 73.00 
            & \textbf{52.79} & \textbf{52.27} & \textbf{71.17} \\

            \midrule
            
            \multicolumn{10}{l}{\textit{\textbf{Qwen3-14B-Base}}} \\
            Baseline
            & 80.80 & 74.15 & 93.60 
            & 45.25 & 40.50 & 65.58 
            & 41.21 & 37.26 & 62.34 \\
            GRPO 
            & 85.00 & 88.35 & 96.40 
            & 58.16 & 58.46 & 74.78 
            & 56.30 & 56.47 & 73.24 \\
            \rowcolor{MintCream}
            \modelname
            & \textbf{90.80} & \textbf{90.80} & \textbf{97.20} 
            & \textbf{61.87} & \textbf{60.90} & \textbf{75.82} 
            & \textbf{61.03} & \textbf{60.07} & \textbf{74.94} \\
            \bottomrule
        \end{tabular}
    }
\end{table*}

To simultaneously encourage solution exploration and consolidate the internalized meta-experiences, achieving dual optimization across trajectory-level behaviors and knowledge-level representations, we train the policy model $\pi_{\theta}$ using a joint optimization objective.
To simultaneously encourage solution exploration and consolidate the internalized meta-experiences, achieving dual optimization across trajectory-level behaviors and knowledge-level representations, we train the policy model $\pi_{\theta}$ using a joint optimization objective.
This objective synergizes the RLVR signal derived from diverse explorative rollouts with the supervised signal distilled from high-quality meta-experiences:
\begin{equation}
    \mathcal{J}(\theta) = \mathcal{J}_{\text{RLVR}}(\theta) + \mathcal{J}_{\text{MEL}}(\theta).
    \label{eq:joint_objective}
\end{equation}

We adopt GRPO~\cite{shao2024deepseekmath} as the RLVR component and compute group-normalized advantages by standardizing rewards within the sampled group and broadcast them to each token.
Let $y_{i,t}$ denote the $t$-th token in trajectory $y_i$ and $y_{i,<t}$, 
the corresponding prefix.
Substituting the definition of $\mathcal{R}_{\text{MEL}}$ from Eq.~\ref{eq:return}, the joint objective in Eq.~\ref{eq:joint_objective} is explicitly expanded as:
\begin{equation}
\small
\begin{aligned}
\mathcal{J}(\theta)
= &\mathbb{E}_{x\sim\mathcal{D},\,\{y_i\}_{i=1}^G\sim\pi_{\theta_{\mathrm{old}}}(\cdot\mid x)} \\
&\Big[ 
\frac{1}{G}\sum_{i=1}^G \frac{1}{|y_i|} \sum_{t=1}^{|y_i|}\min\Big( \rho_{i,t}(\theta)\hat{A}_{i,t},\; \\
&\mathrm{clip}\big(\rho_{i,t}(\theta), 1-\epsilon, 1+\epsilon\big)\hat{A}_{i,t} \Big) +  \mathcal{R}_{\text{MEL}}\Big].
\end{aligned}
\end{equation}

Although derived from a log-likelihood objective, its optimization gradient is mathematically equivalent to a policy gradient update where the reward signal is a constant positive scalar.
Consequently, the total objective $\mathcal{J}(\theta)$ can be unified as maximizing the expected cumulative return of a hybrid reward function. In this unified view, the meta-experiences function as a dense process reward model.

Unlike the sparse outcome rewards in standard RLVR that only evaluate the final correctness, $\mathcal{R}_{\text{MEL}}$ provides explicit, step-by-step reinforcement for the reasoning process itself.
This ensures that the model not only pursues correct outcomes via broad exploration but is also continuously shaped by the dense supervision of its own successful cognitive patterns, effectively bridging the gap between trajectory-level search and token-level knowledge encoding.
\section{Experiments}

\paragraph{Datasets.}  
We train our model on the DAPO-Math-17k dataset~\cite{yu2025dapo} and evaluate it on five challenging mathematical reasoning benchmarks: AIME24, AIME25, AMC23~\cite{li2024numinamath}, MATH500~\cite{math500}, and OlympiadBench~\cite{he2024olympiadbench}. 

\paragraph{Setups.}
All reinforcement learning training is conducted using the VERL framework~\cite{sheng2024hybridflow} on 8$\times$H20 GPUs, with Math-Verify providing rule-based outcome verification. During training, we sample 8 responses per prompt at a temperature of 1.0 with a batch size of 128. Optimization uses a learning rate of $1\times10^{-6}$ and a mini-batch size of 128.
For evaluation, we report Pass@1 at temperature 0, and Avg@8 and Pass@8 at temperature 0.6.

\paragraph{Models and Baselines.}
To demonstrate the general applicability of \modelname, we conduct experiments across a diverse range of model scales, including Qwen3-4B-Base, Qwen3-8B-Base, and Qwen3-14B-Base~\cite{yang2025qwen3}.
We adopt GRPO~\cite{shao2024deepseekmath} as the base reinforcement learning algorithm for \modelname, and thus perform a direct and controlled comparison between the vanilla GRPO and our meta-experience learning approach.

\subsection{Experimental Results}
As shown in Table~\ref{tab:main_results_avg8}, \modelname achieves consistent and significant improvements over vanilla GRPO and the basemodel across multiple benchmarks and model scales.
We report three complementary metrics: Pass@1 reflects one-shot reliability, Avg@8 measures the average performance over 8 samples, and Pass@8 reports the best-of-8 success rate.

First, the gains in Pass@1 demonstrate that \modelname substantially enhances the model's confidence in following correct reasoning paths. Across all model scales, it achieves a consistent improvement of 3.92–4.73\% over the strong GRPO baseline. This indicates that \modelname effectively internalizes the explored insights into the model's parametric memory. By consolidating these successful reasoning patterns, the model generates high-confidence solutions, markedly reducing the need for extensive test-time sampling.
This reliability is further corroborated by the gains in Avg@8, which reveal that \modelname significantly enhances reasoning consistency and output stability.
High performance on this metric supports our hypothesis that internalized meta-experiences function as intrinsic process-level guidance, continuously steering the generation toward valid logic and effectively reducing variance across sampled outputs.
Finally, the sustained gains in Pass@8 suggest that learning from meta-experience does not harm exploration; instead, it expands the reachable solution space and raises the upper bound of best-of-$k$ performance.

\subsection{Training Dynamics and Convergence Analysis}

\begin{figure}[ht]
\centering
\centerline{\includegraphics[width=1\columnwidth]{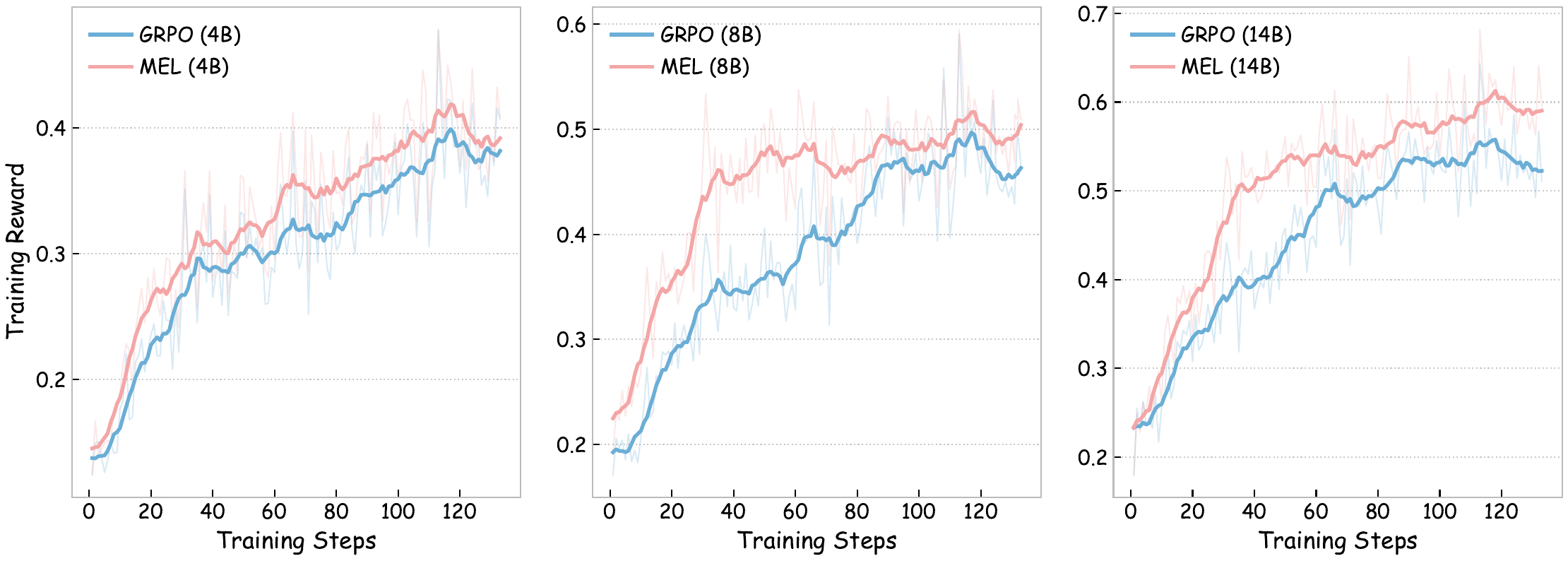}}
    \caption{Training curves comparing GRPO and \modelname.}
    \label{fig:training_curve}
\end{figure}

To understand the mechanisms driving the performance gains under \modelname, we monitored the training dynamics and validation performance in Figures~\ref{fig:training_curve} and \ref{fig:4b_performance_evolution}--\ref{fig:14b_performance_evolution}.

Vanilla GRPO methods often struggle to obtain positive reinforcement in the early stages, particularly when initial performance is low, due to the sparsity of outcome-based rewards.
As illustrated in the training curve, vanilla GRPO exhibits a relatively slow ascent during the initial phase.
In contrast, \modelname demonstrates a sharp, rapid trajectory growth immediately from the onset of training.
This acceleration is attributed to the internalized meta-experience return, $\mathcal{R}_{\text{MEL}}$.
By functioning as a dense, language-modeling process reward, $\mathcal{R}_{\text{MEL}}$ continuously provides informative gradient signals for every reasoning step, even when successful trajectories yielding positive reinforcement are scarce.

Beyond sample efficiency, \modelname achieves a consistently higher performance upper bound.
The training curves show that the average reward of \modelname consistently surpasses that of vanilla GRPO throughout the entire training process.
Crucially, the downstream validation trajectories reveal that even as performance growth begins to plateau in the later stages, \modelname maintains a distinct and sustained advantage over the baseline.
This phenomenon demonstrates that the internalization of meta-experiences empowers the model to effectively navigate and explore more complex, long-horizon solutions that remain inaccessible to the baseline.

\begin{figure*}[t]
\centering
\centerline{\includegraphics[width=0.99\textwidth]{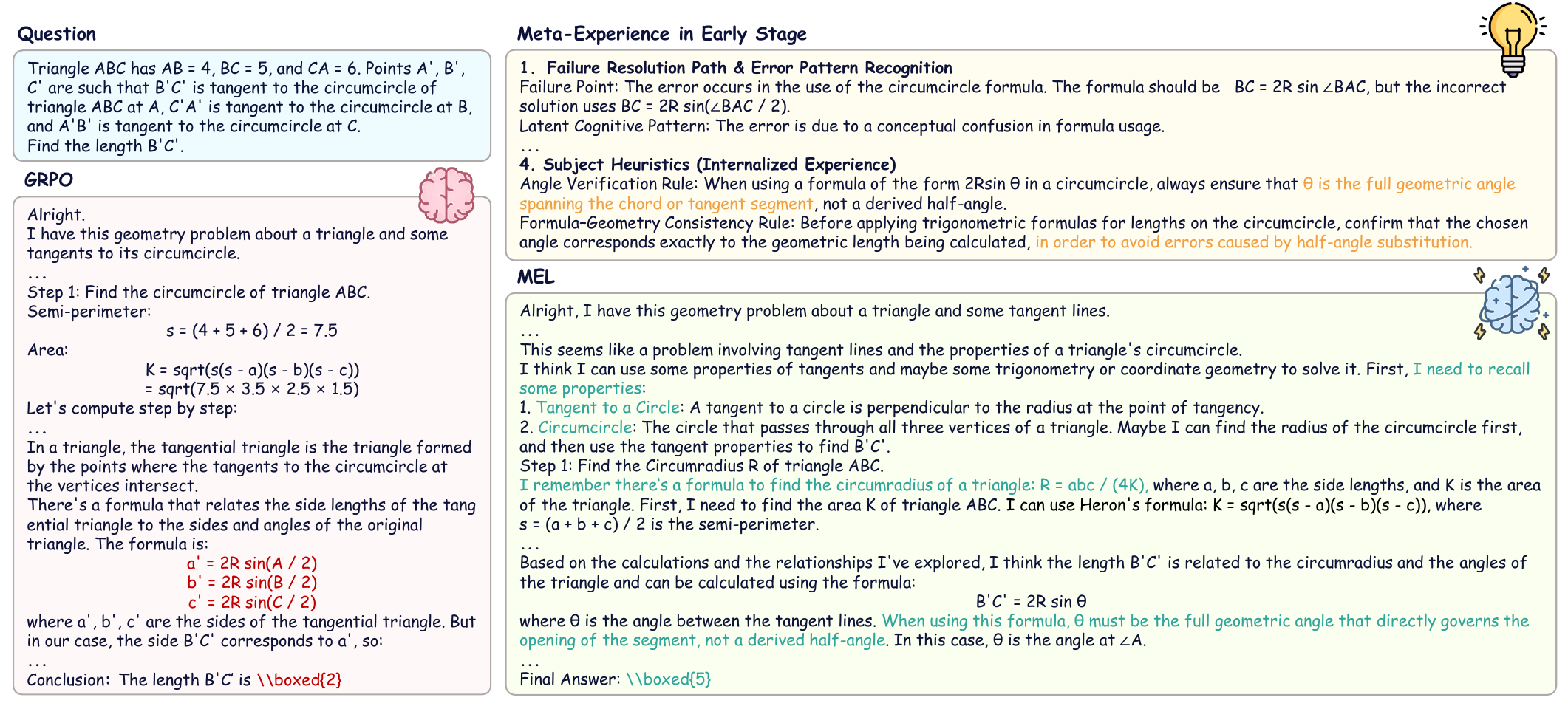}}
    \caption{Case study comparing GRPO and \modelname, with visualization of meta-experience in early stage.}
    \label{fig:case_study}
\end{figure*}

\subsection{How Meta-Experience Shapes Reasoning Patterns}
To investigate how \modelname shapes the model's cognitive processes beyond numerical metrics, we conduct a qualitative analysis comparing the reasoning trajectories of \modelname and the baseline GRPO model, as visualized in Figure~\ref{fig:case_study}.

A distinct behavioral divergence is observed from the onset of the solution.
While the GRPO baseline tends to prioritize immediate execution through direct numerical operations, \modelname adopts a structured preparatory strategy by explicitly outlining relevant theorems and formulas.
Although the direct approach may appear efficient for simple queries, it increases the susceptibility to errors in complex tasks due to the lack of a holistic view of problem constraints.

Notably, \modelname exhibits an emergent cognitive behavior.
When applying specific theorems, it spontaneously activates internalized ``bitter lessons'' as endogenous safeguards to regulate its actions.
These active signals effectively reduce reasoning drift by encouraging earlier constraint checking and consistent self-correction when the model enters uncertain regions.

\subsection{Generality Across Learning Paradigms}
\begin{figure}[ht]
  \centering
  \begin{minipage}{0.235\textwidth}
    \centering
    \includegraphics[width=\textwidth]{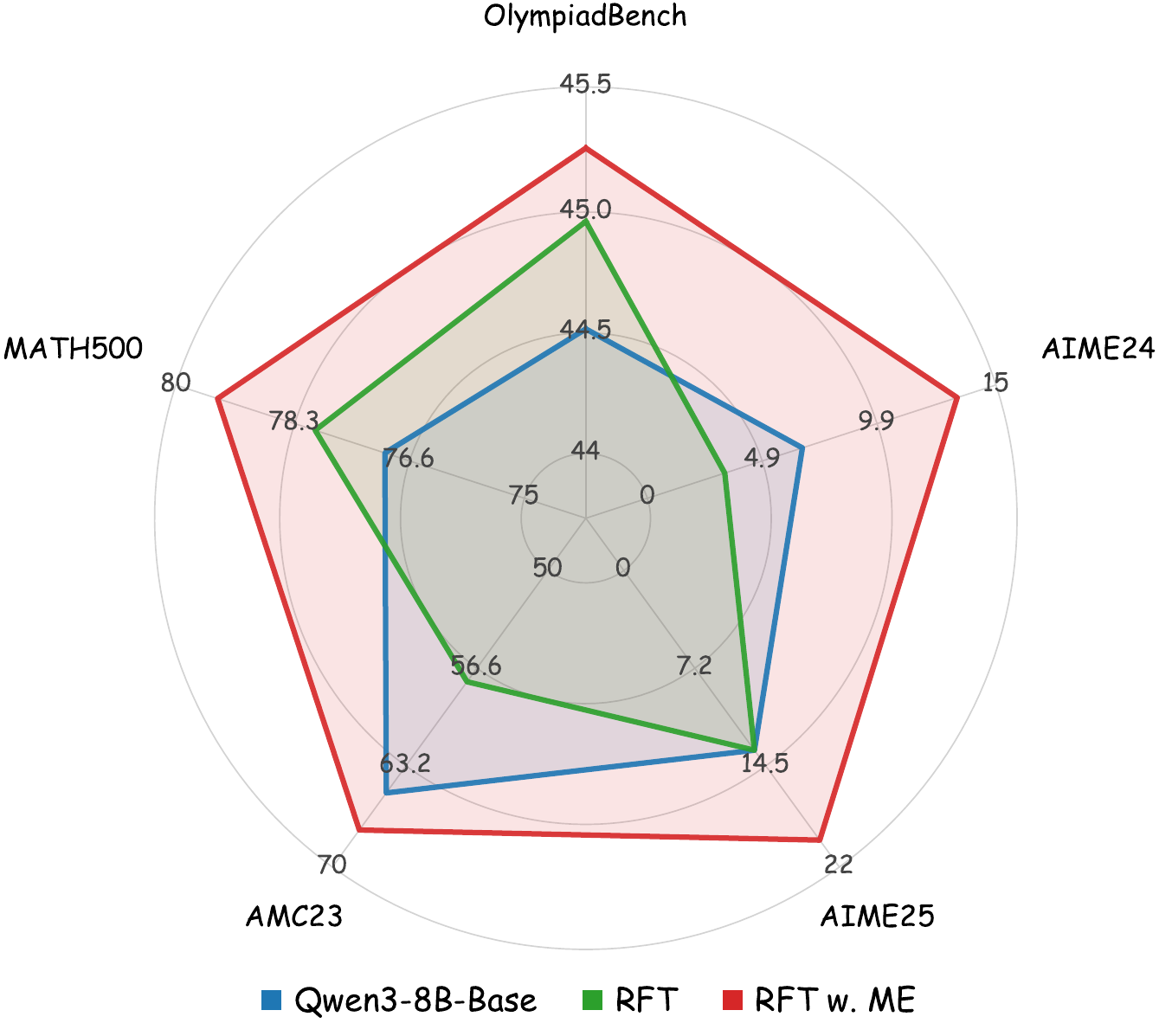}
  \end{minipage}
  \hfill
  \begin{minipage}{0.235\textwidth}
    \centering
    \includegraphics[width=\textwidth]{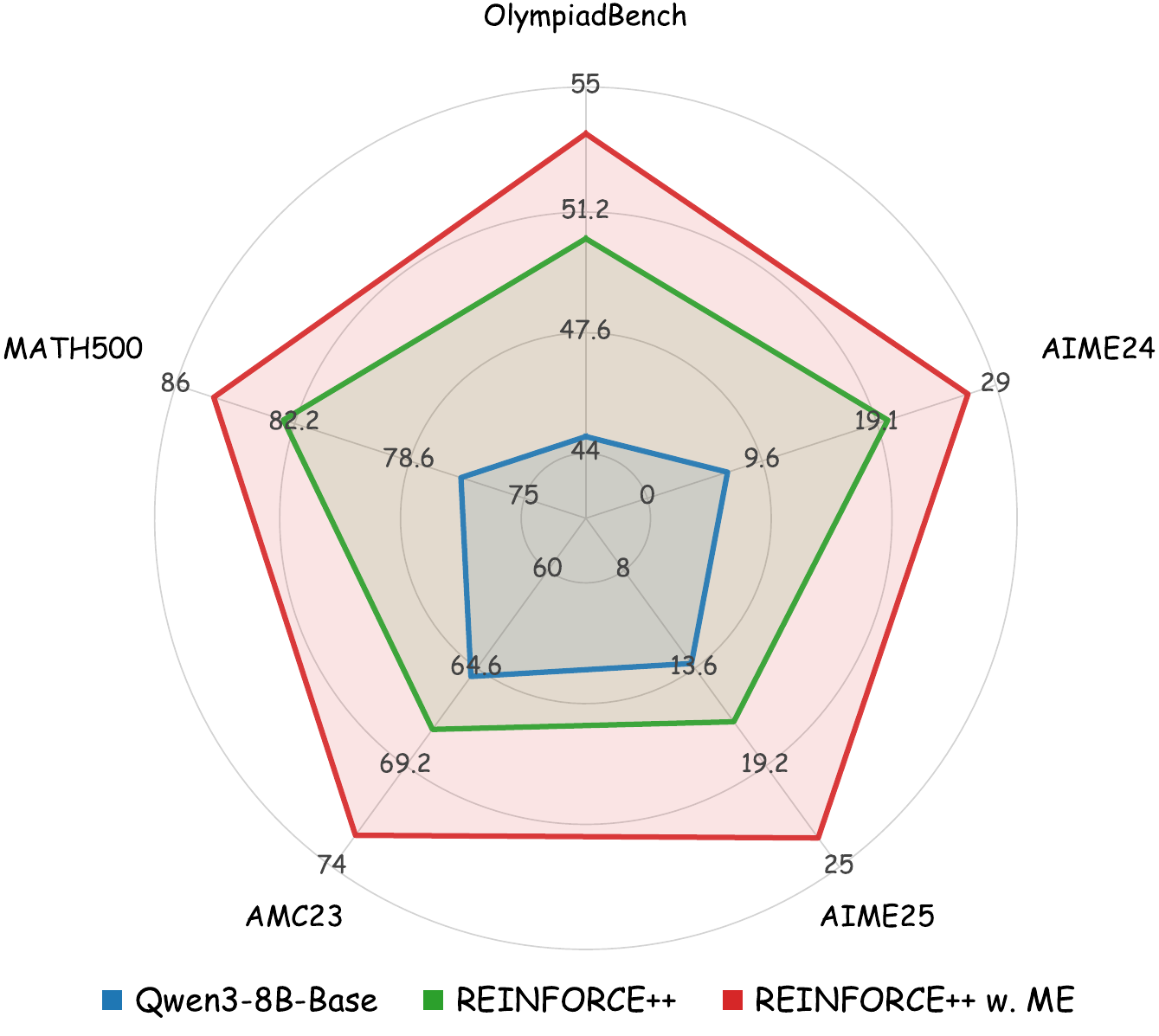}
  \end{minipage}
  \caption{Impact of meta-experience across different training methods, including Rejection Sampling Fine-Tuning (RFT) and REINFORCE++. ME denotes \textbf{Meta-Experience}.}
  \label{fig:radar_performance}
\end{figure}
To demonstrate the versatility of meta-experience, we integrated it into RFT and REINFORCE++ using the Qwen-8B-Base model as the backbone and the same training set in our experiments. As shown in Figure~\ref{fig:radar_performance}, while vanilla RFT often suffers from rote memorization and tends to overfit to specific samples in this training set, the incorporation of meta-experiences introduces robust reasoning heuristics. This allows the model to internalize the underlying logic rather than merely imitating specific answers, thereby effectively mitigating overfitting and enhancing generalization to unseen test sets. Similarly, applying meta-experiences to REINFORCE++ significantly raises the performance ceiling on benchmarks. This confirms that the benefit of internalized meta-experiences is a universal enhancement, not limited to the GRPO framework.

\subsection{Scalability Analysis}
As indicated by the training curves in Figure~\ref{fig:training_curve}, the method exhibits a distinct positive scaling law: the performance margin between \modelname and the baseline widens significantly as the model size increases. 
This phenomenon consistently extends to downstream validation benchmarks.

We attribute this effect to the quality of self-generated supervision, which is inherently bounded by the model’s intrinsic capability. As shown in Figure~\ref{fig:retention_ratio}, the 14B model achieves a significantly higher yield rate of valid meta-experiences than its smaller counterparts. While limited-capacity models introduce noise due to imprecise error attribution, larger models benefit from stronger self-verification, enabling the distillation of high-quality heuristics that provide more accurate gradient signals and fully realize the potential of our framework.

\section{Conclusion}
In this paper, we introduced \textbf{\modelname}, a novel framework designed to overcome the meta-learning bottleneck in standard RLVR by transforming instance-specific failure patterns into reusable cognitive assets.
Unlike traditional methods that rely solely on outcome-oriented rewards, \modelname empowers models to perform granular error attribution, distilling specific failure modes into natural language heuristics—termed \textbf{Meta-Experiences}.
By internalizing these experiences into parametric memory, our approach bridges the gap between verifying a solution and understanding the underlying reasoning logic.
Extensive empirical evaluations confirm that \modelname consistently boosts mathematical reasoning across diverse model scales.

\section*{Impact Statement}
This paper presents research aimed at advancing the field of reinforcement learning. While our work may have broader societal implications, we do not identify any specific impacts that require particular attention at this stage.

\bibliography{icml}
\bibliographystyle{icml2026}

\newpage
\appendix
\onecolumn
\clearpage

\section{Result of Performance Evolution}
\begin{figure*}[ht]
  \centering
  \begin{minipage}{0.32\textwidth}
    \centering
    \includegraphics[width=\textwidth]{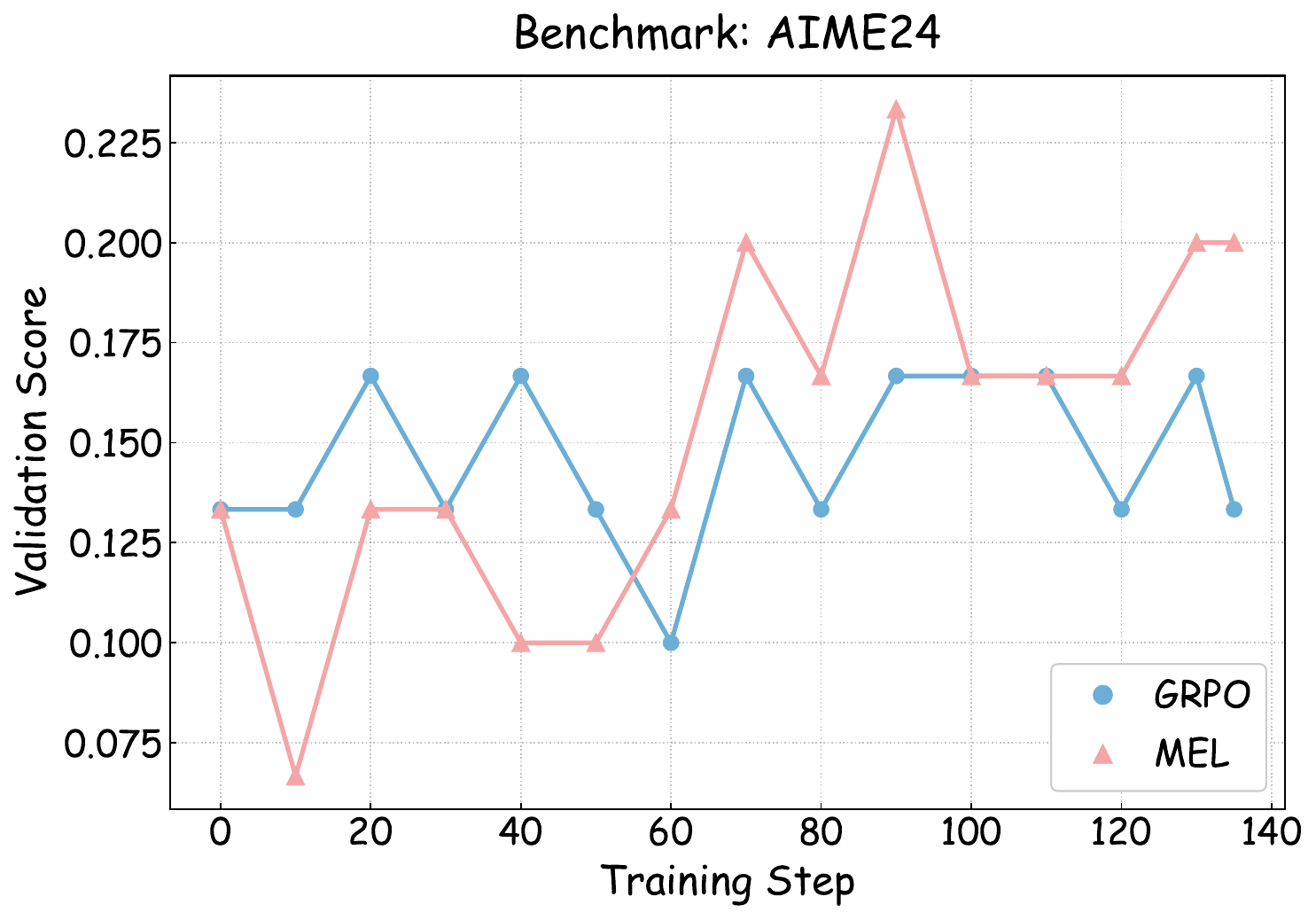}
  \end{minipage}
  \hfill
  \begin{minipage}{0.32\textwidth}
    \centering
    \includegraphics[width=\textwidth]{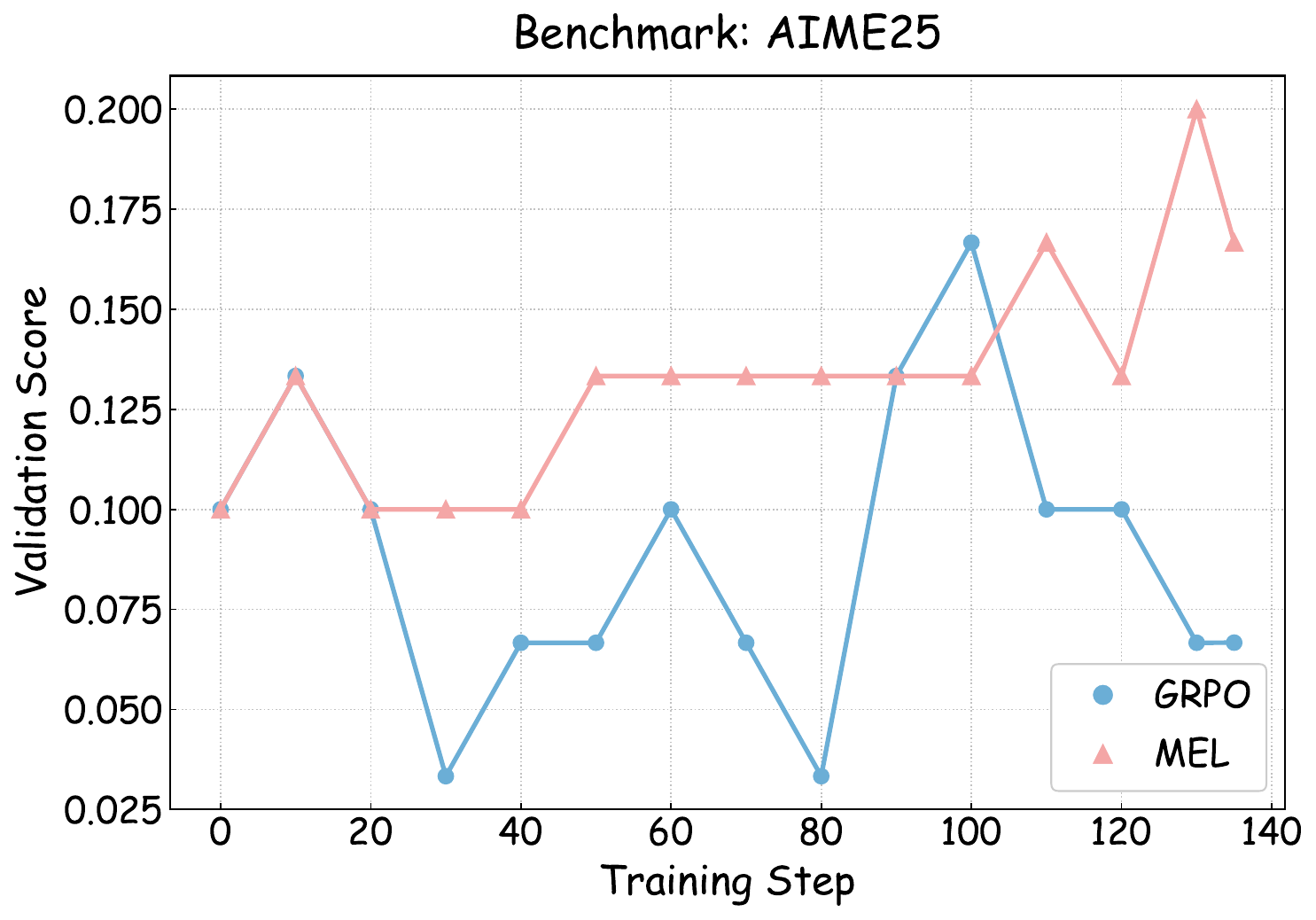}
  \end{minipage}
  \hfill
  \begin{minipage}{0.32\textwidth}
    \centering
    \includegraphics[width=\textwidth]{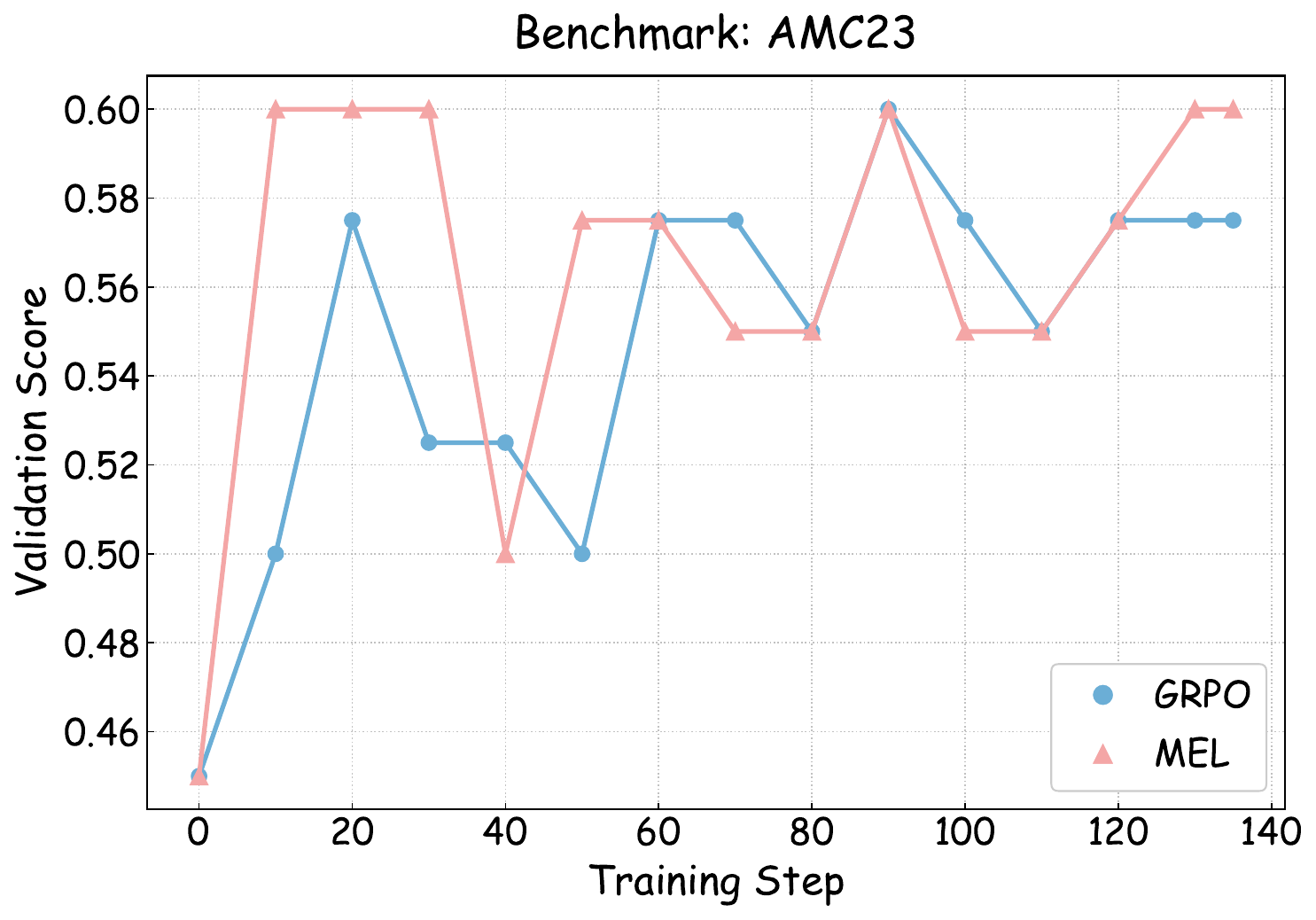}
  \end{minipage}

  \vspace{0.15cm}

  \begin{minipage}{0.32\textwidth}
    \centering
    \includegraphics[width=\textwidth]{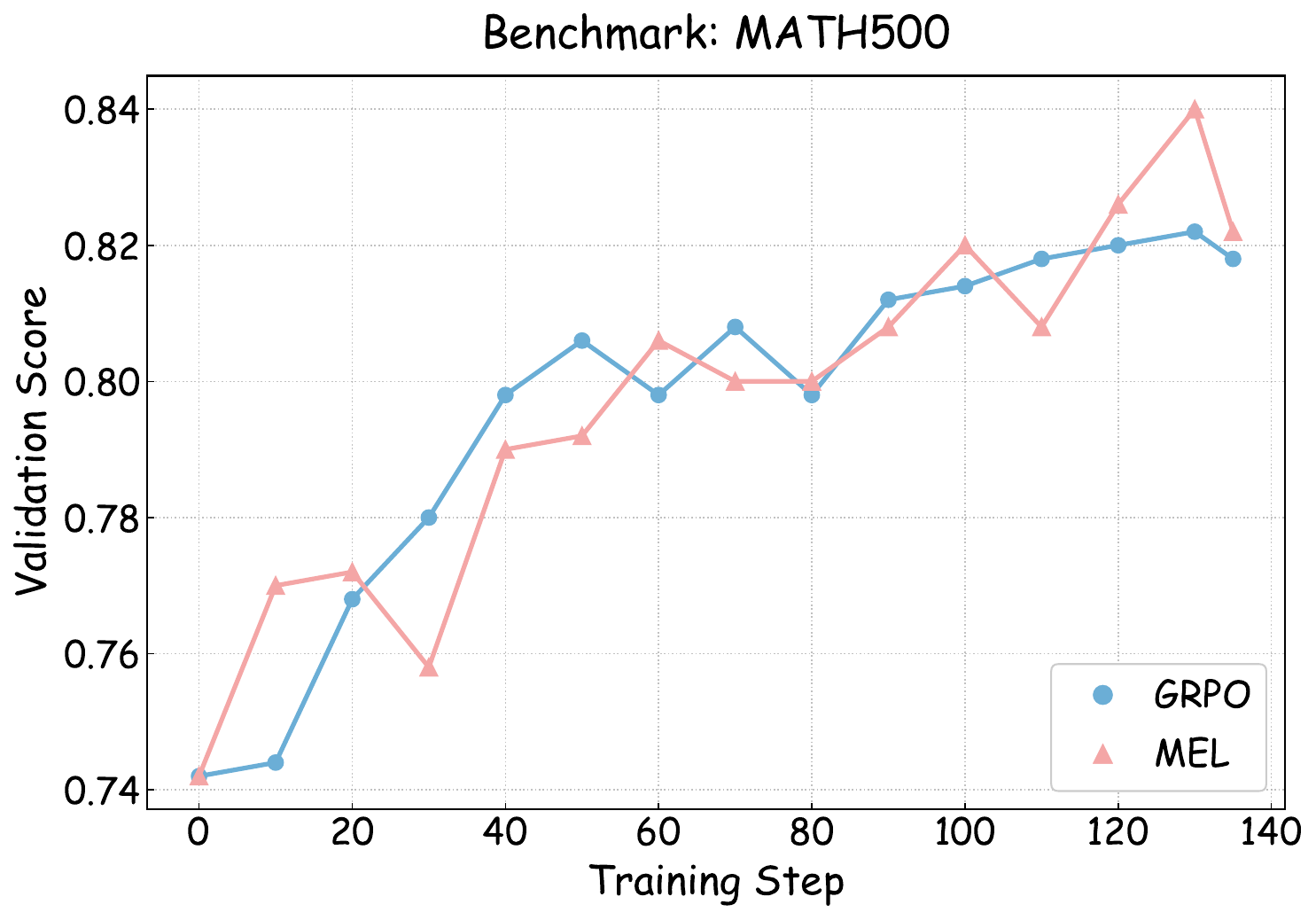}
  \end{minipage}
  \hfill
  \begin{minipage}{0.32\textwidth}
    \centering
    \includegraphics[width=\textwidth]{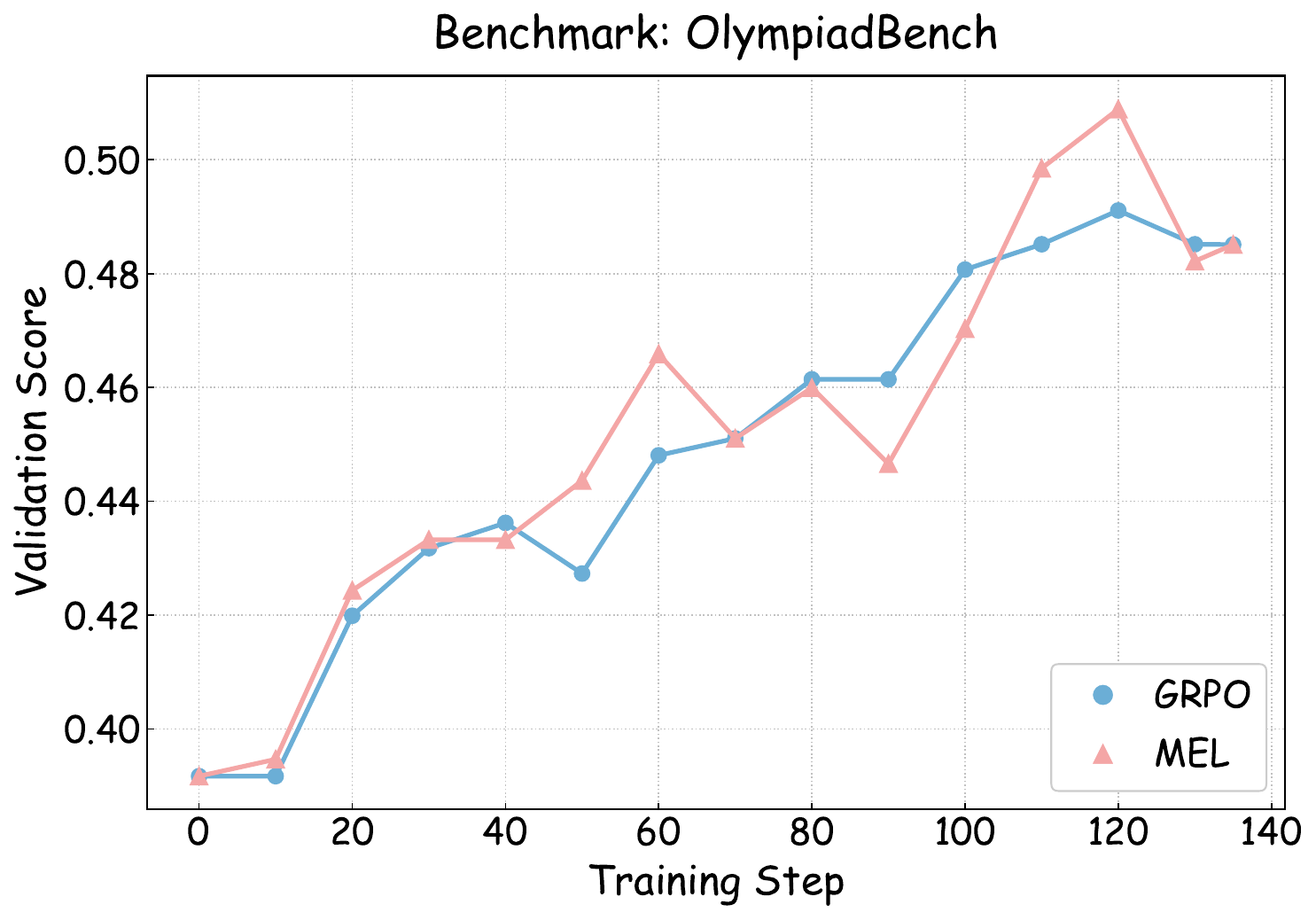}
  \end{minipage}
  \hfill
  \begin{minipage}{0.32\textwidth}
    \centering
    \includegraphics[width=\textwidth]{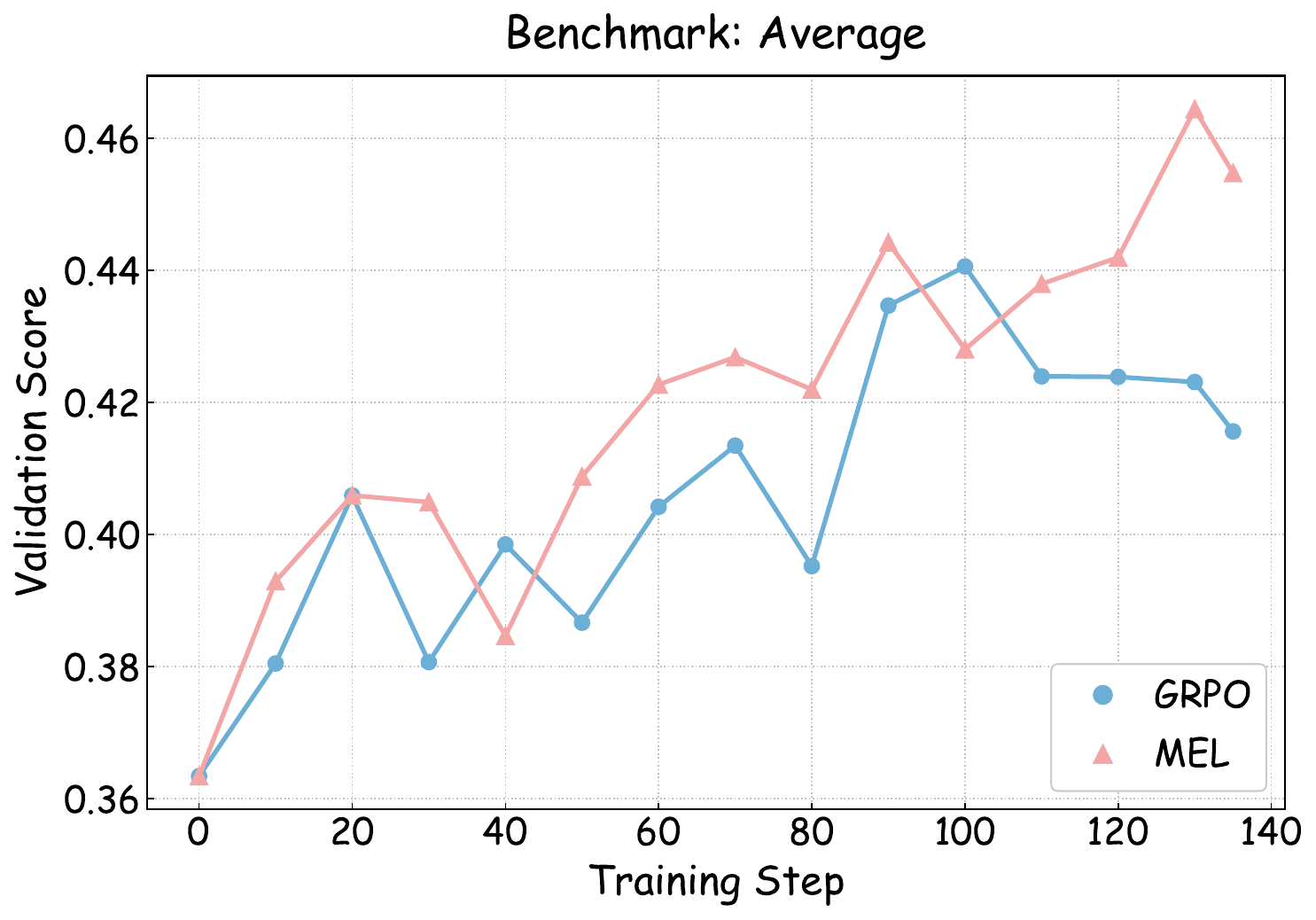}
  \end{minipage}

  \caption{Performance evolution of GRPO and \modelname on Qwen3-4B-Base across training steps on multiple benchmarks.}
  \label{fig:4b_performance_evolution}
\end{figure*}

\begin{figure*}[ht]
  \centering
  \begin{minipage}{0.32\textwidth}
    \centering
    \includegraphics[width=\textwidth]{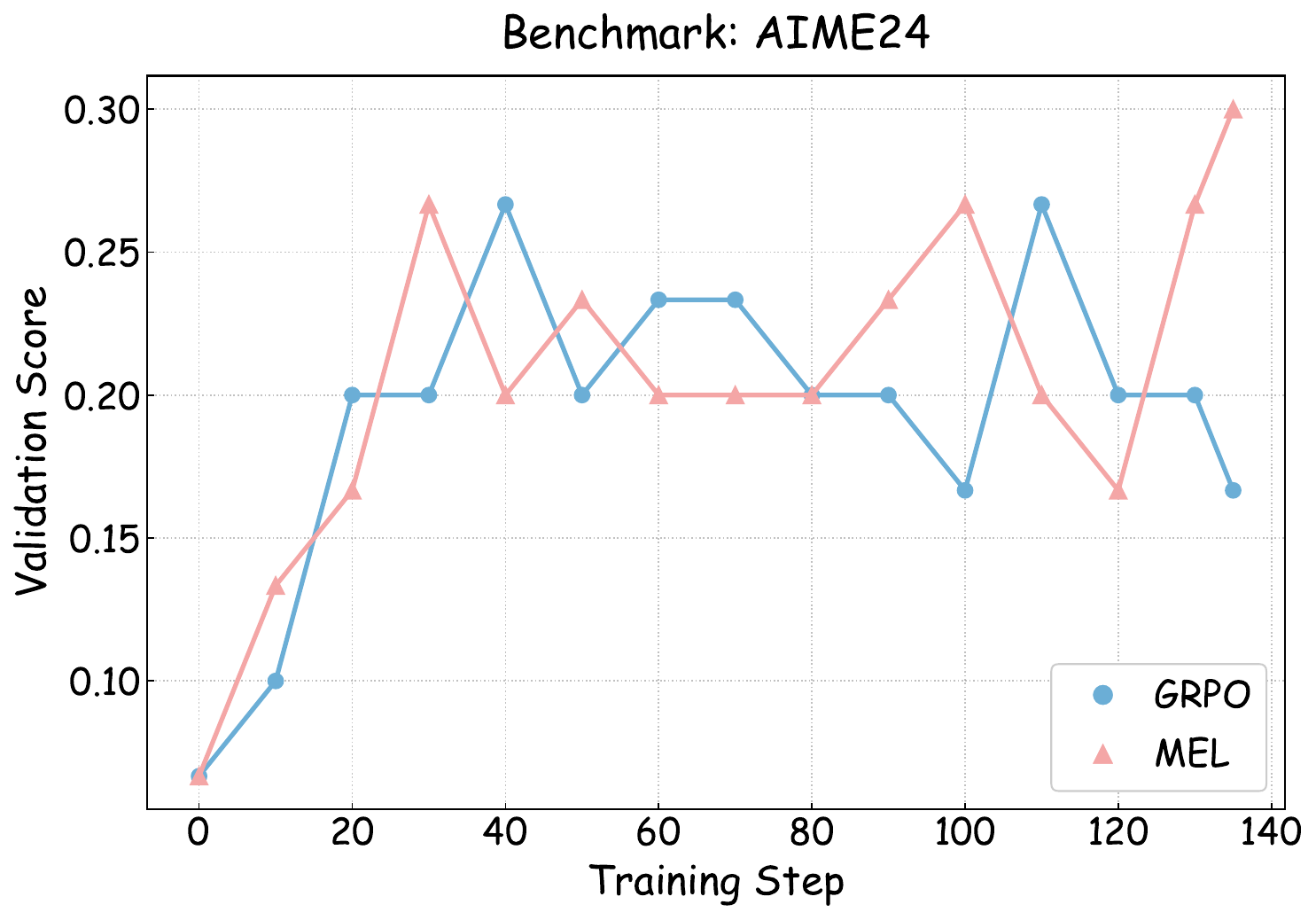}
  \end{minipage}
  \hfill
  \begin{minipage}{0.32\textwidth}
    \centering
    \includegraphics[width=\textwidth]{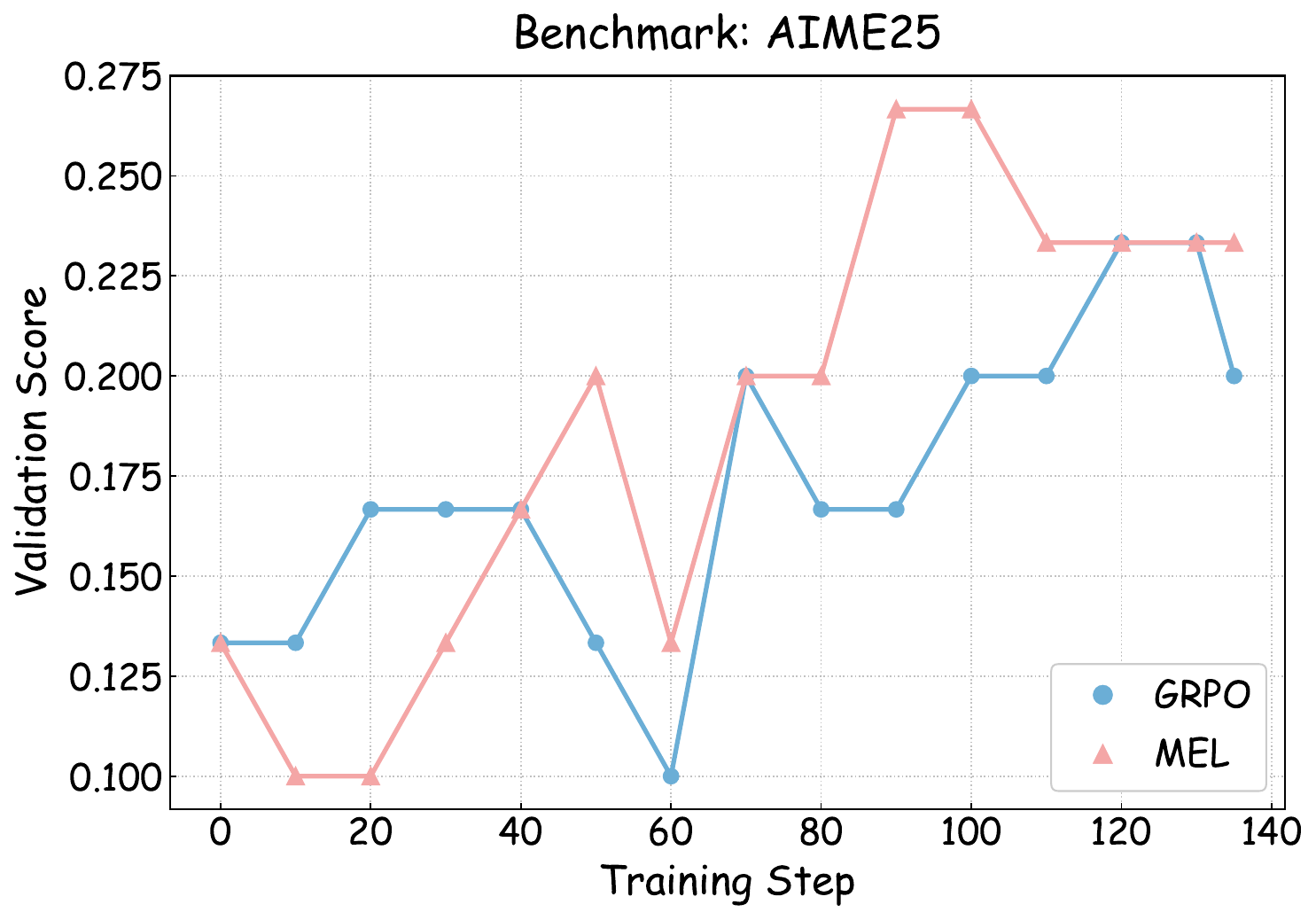}
  \end{minipage}
  \hfill
  \begin{minipage}{0.32\textwidth}
    \centering
    \includegraphics[width=\textwidth]{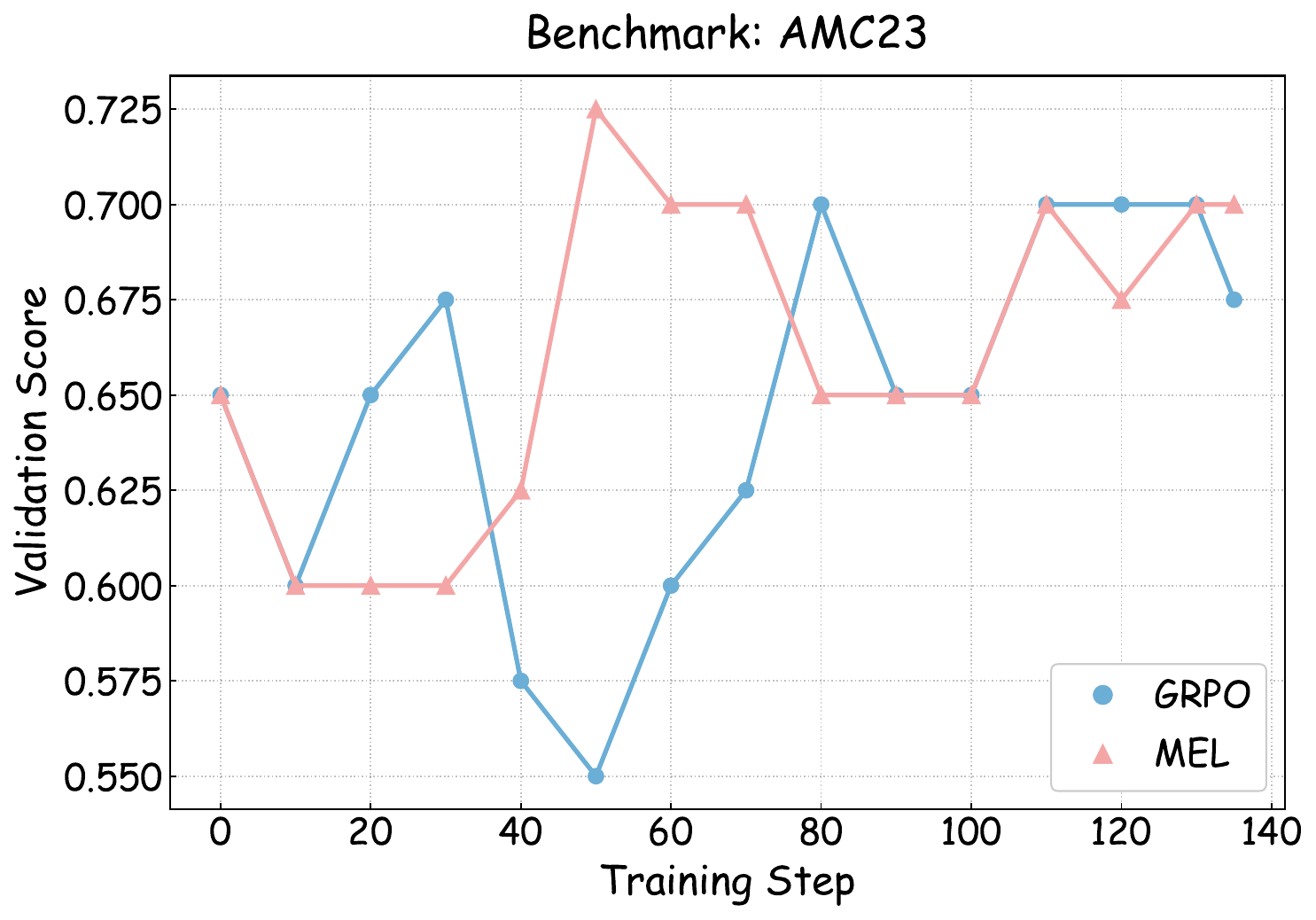}
  \end{minipage}

  \vspace{0.15cm}

  \begin{minipage}{0.32\textwidth}
    \centering
    \includegraphics[width=\textwidth]{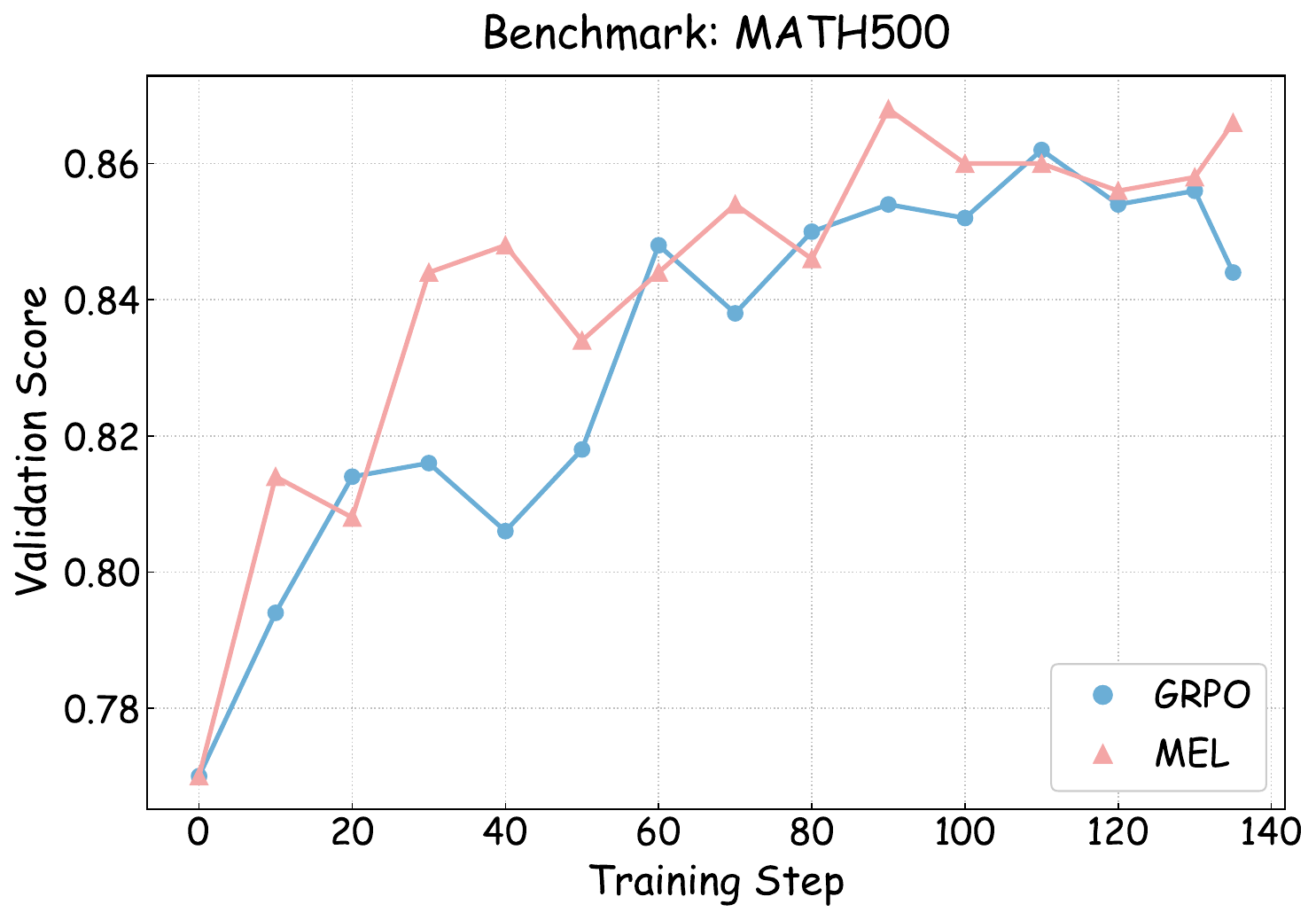}
  \end{minipage}
  \hfill
  \begin{minipage}{0.32\textwidth}
    \centering
    \includegraphics[width=\textwidth]{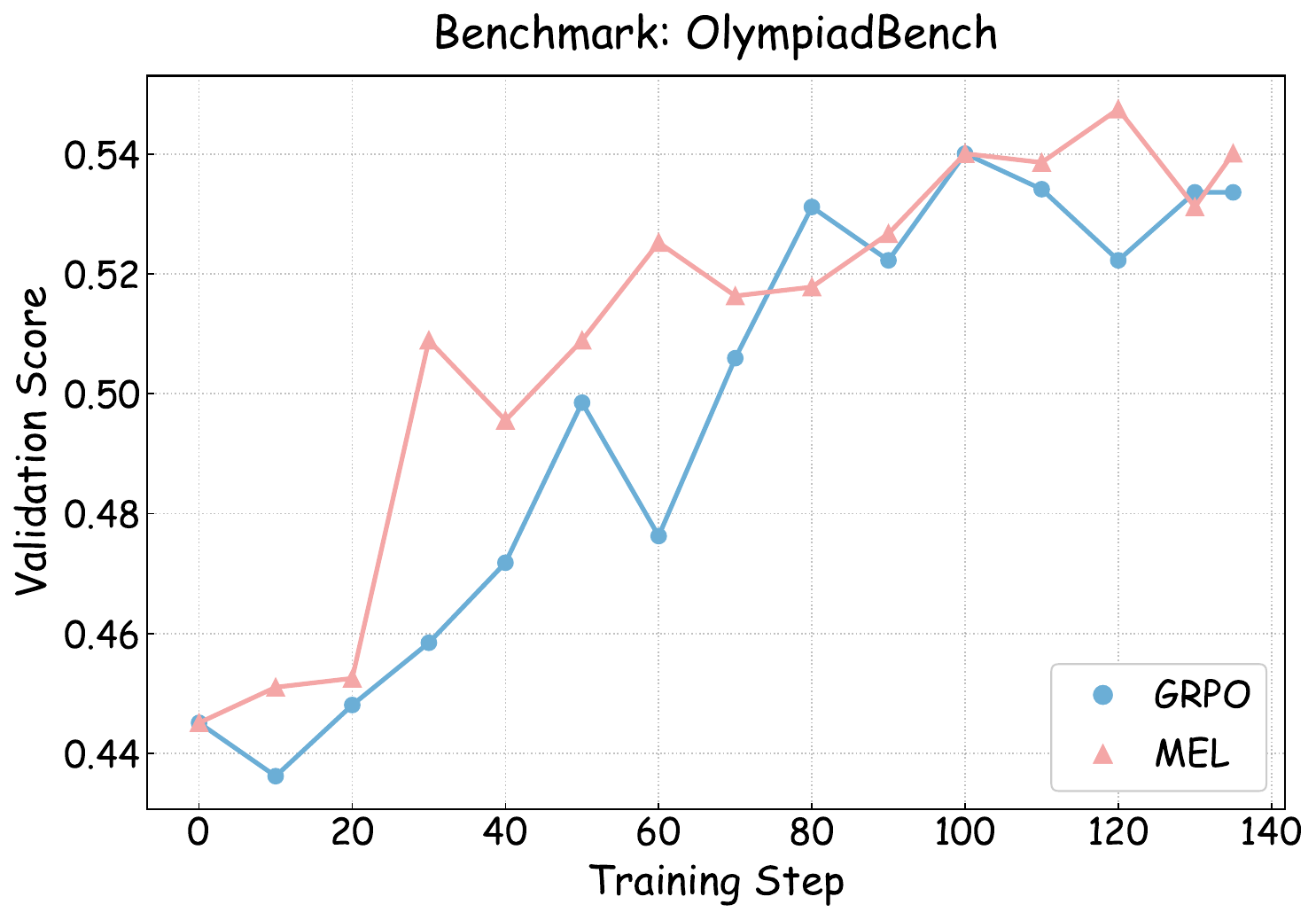}
  \end{minipage}
  \hfill
  \begin{minipage}{0.32\textwidth}
    \centering
    \includegraphics[width=\textwidth]{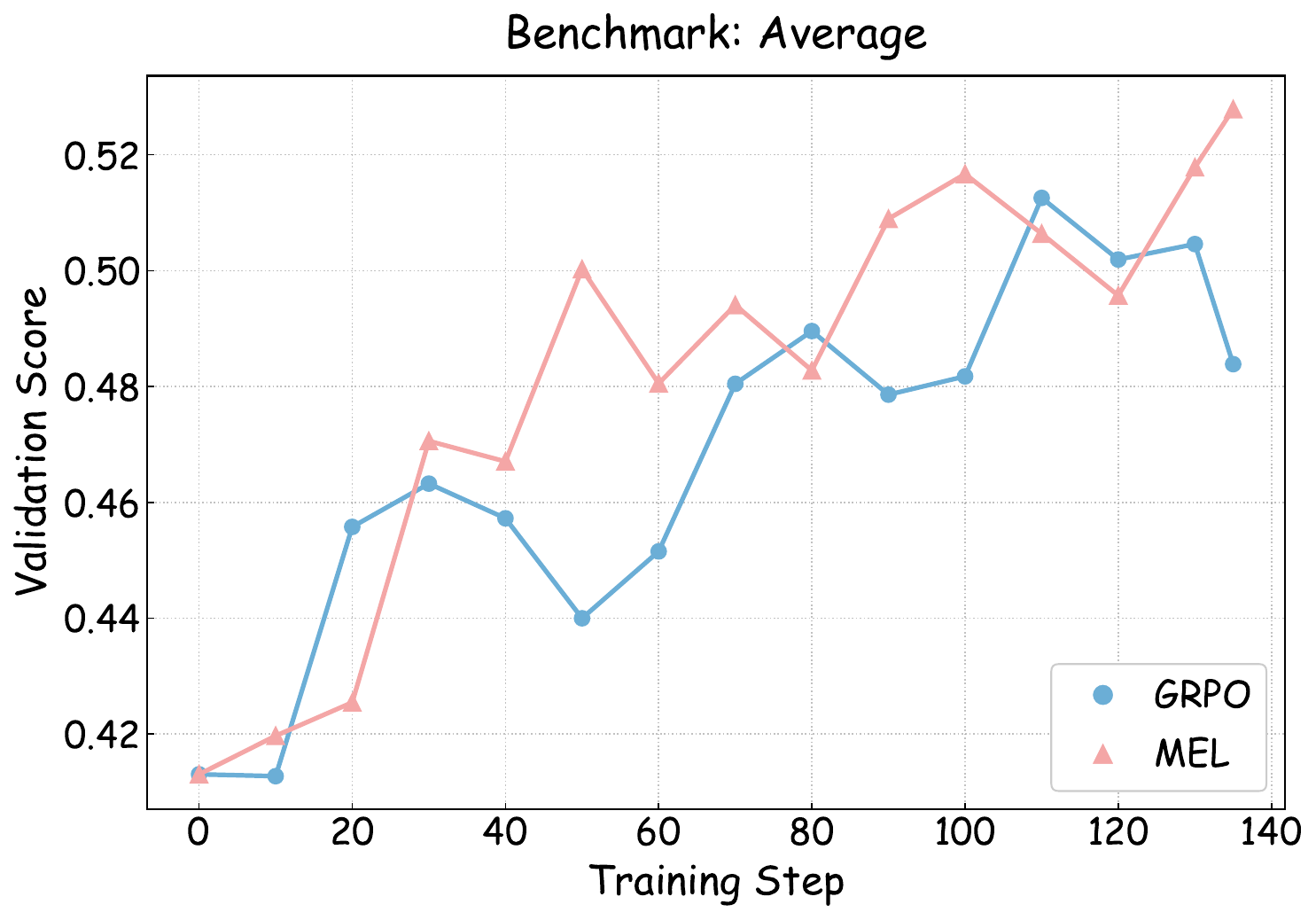}
  \end{minipage}

  \caption{Performance evolution of GRPO and \modelname on Qwen3-8B-Base across training steps on multiple benchmarks.}
  \label{fig:8b_performance_evolution}
\end{figure*}

\begin{figure*}[h]
  \centering
  \begin{minipage}{0.32\textwidth}
    \centering
    \includegraphics[width=\textwidth]{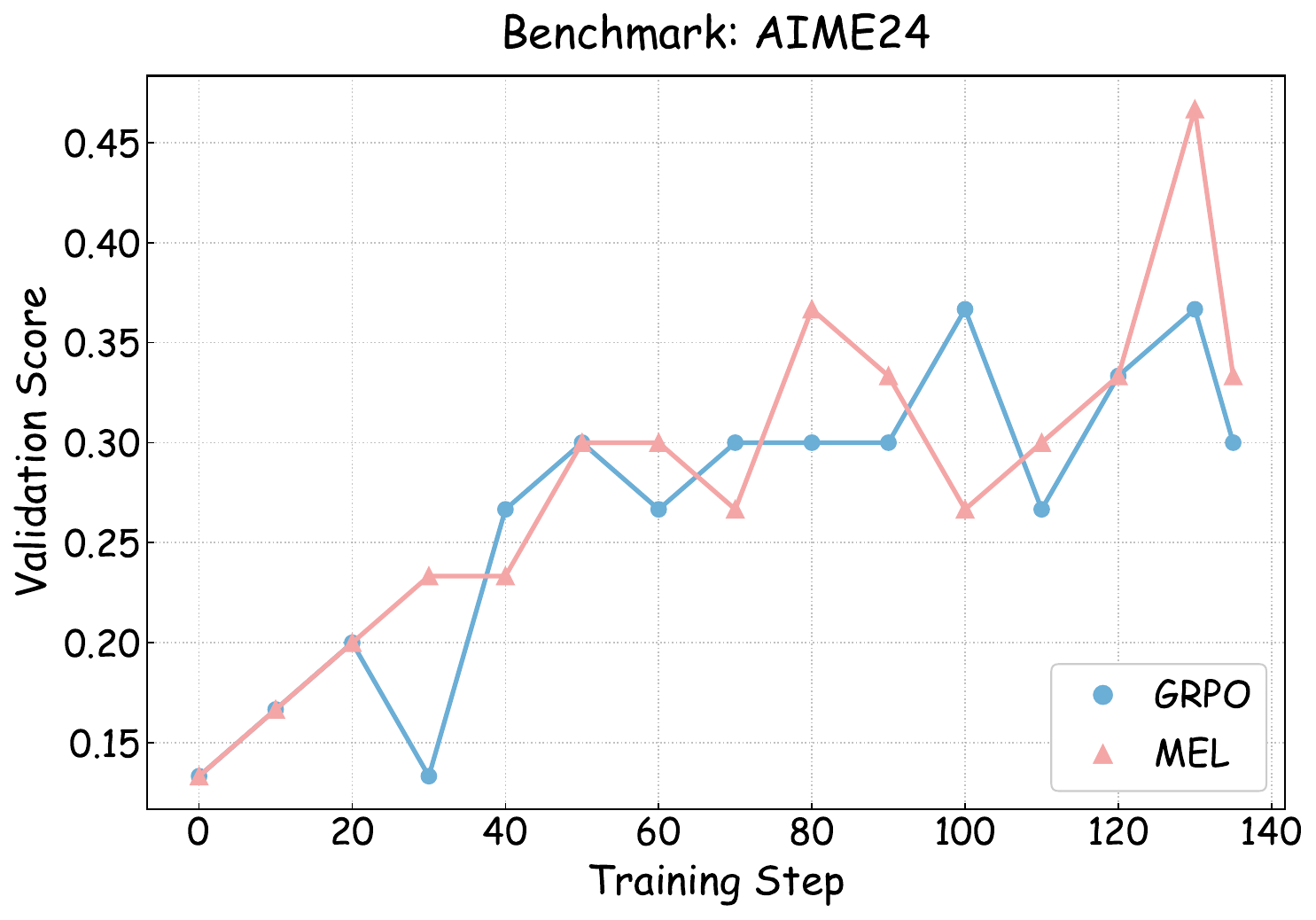}
  \end{minipage}
  \hfill
  \begin{minipage}{0.32\textwidth}
    \centering
    \includegraphics[width=\textwidth]{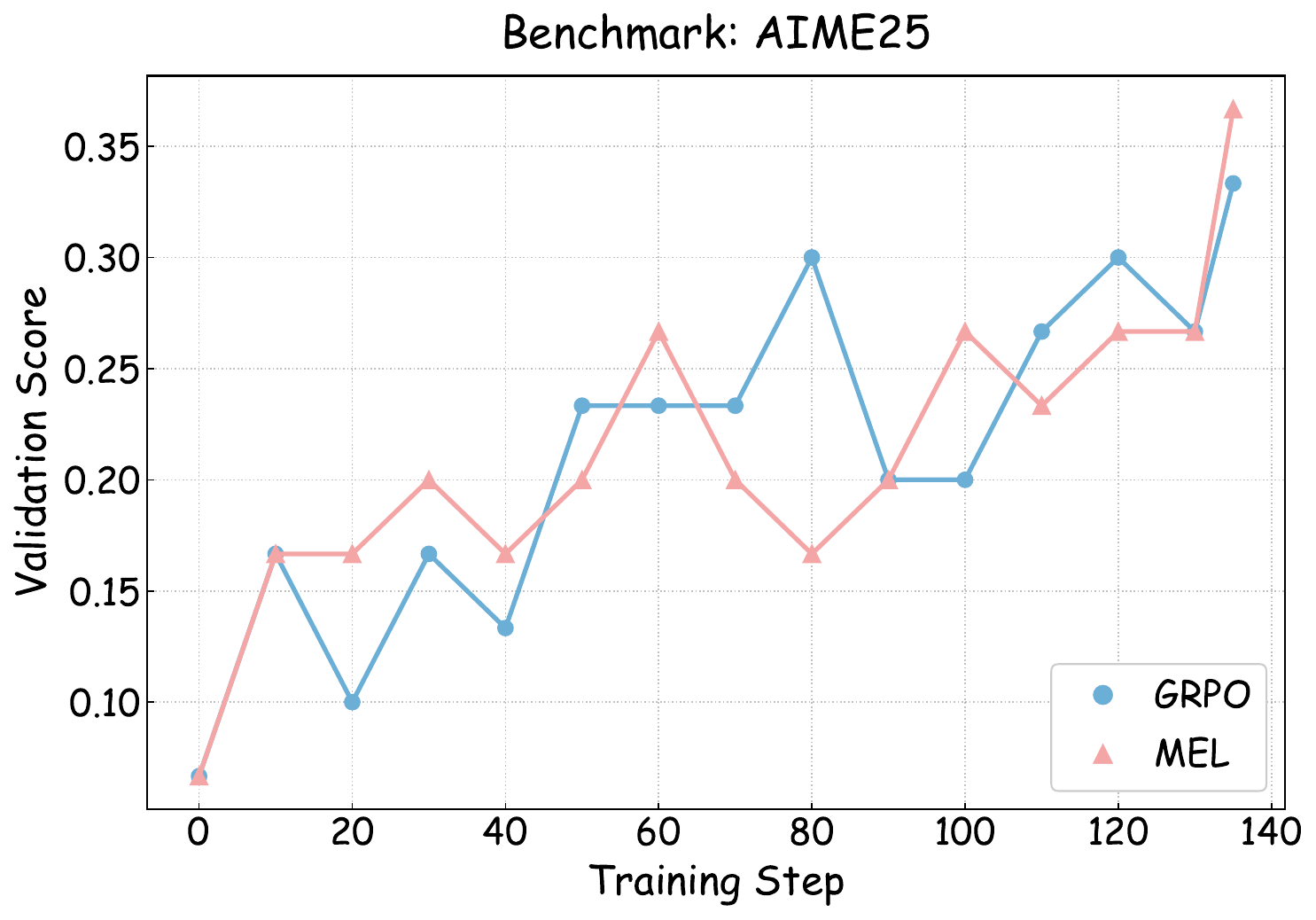}
  \end{minipage}
  \hfill
  \begin{minipage}{0.32\textwidth}
    \centering
    \includegraphics[width=\textwidth]{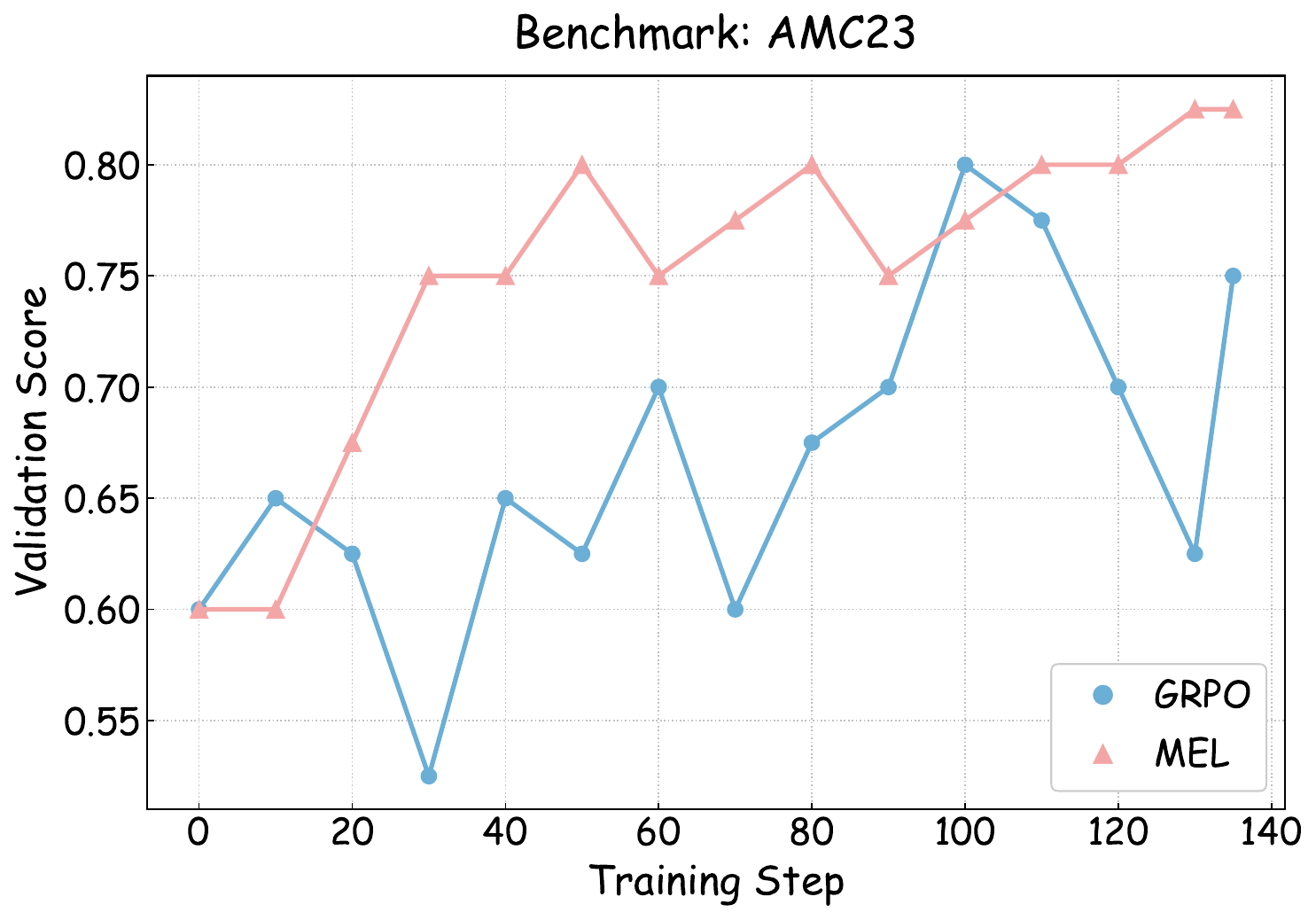}
  \end{minipage}

  \vspace{0.15cm}

  \begin{minipage}{0.32\textwidth}
    \centering
    \includegraphics[width=\textwidth]{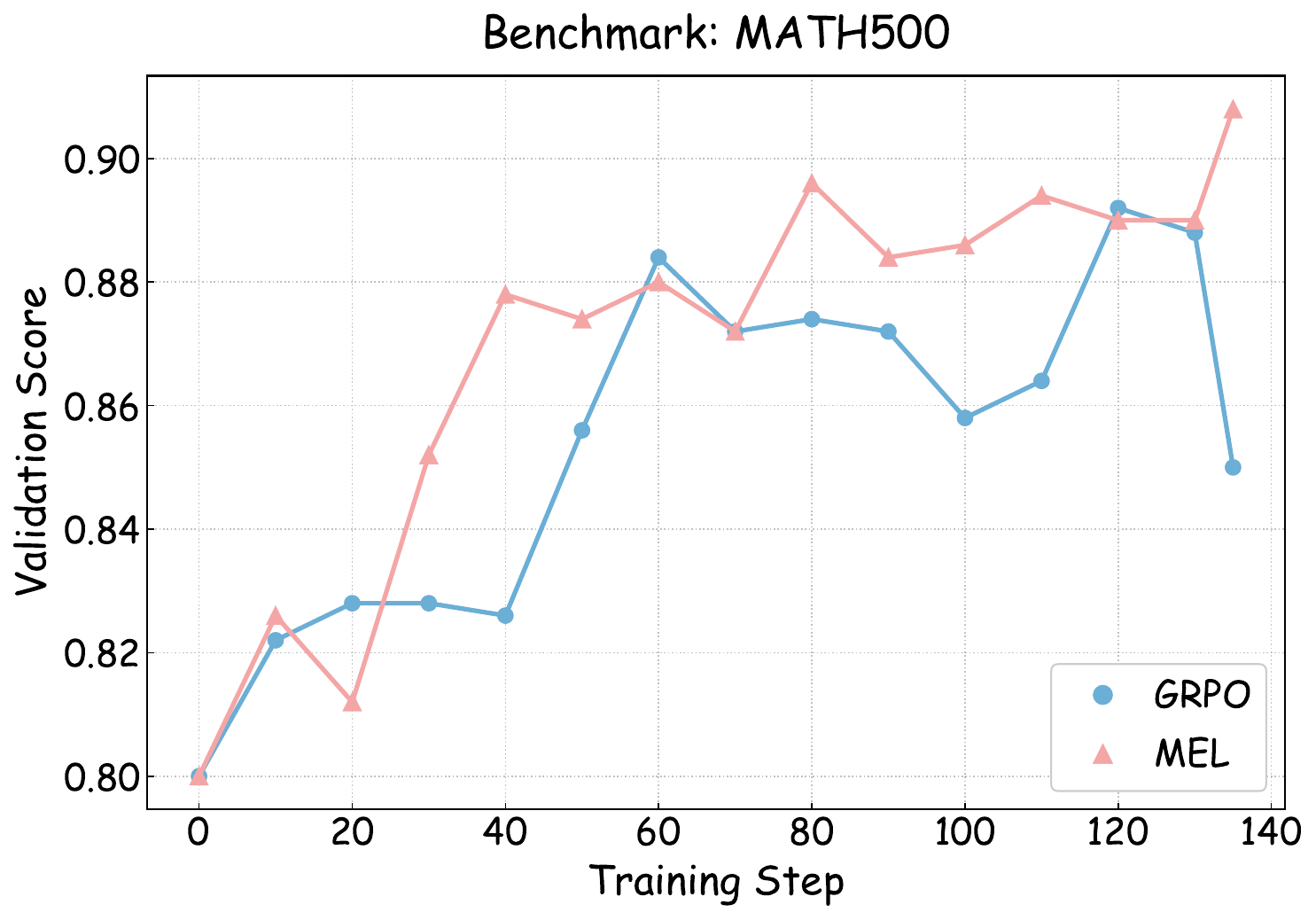}
  \end{minipage}
  \hfill
  \begin{minipage}{0.32\textwidth}
    \centering
    \includegraphics[width=\textwidth]{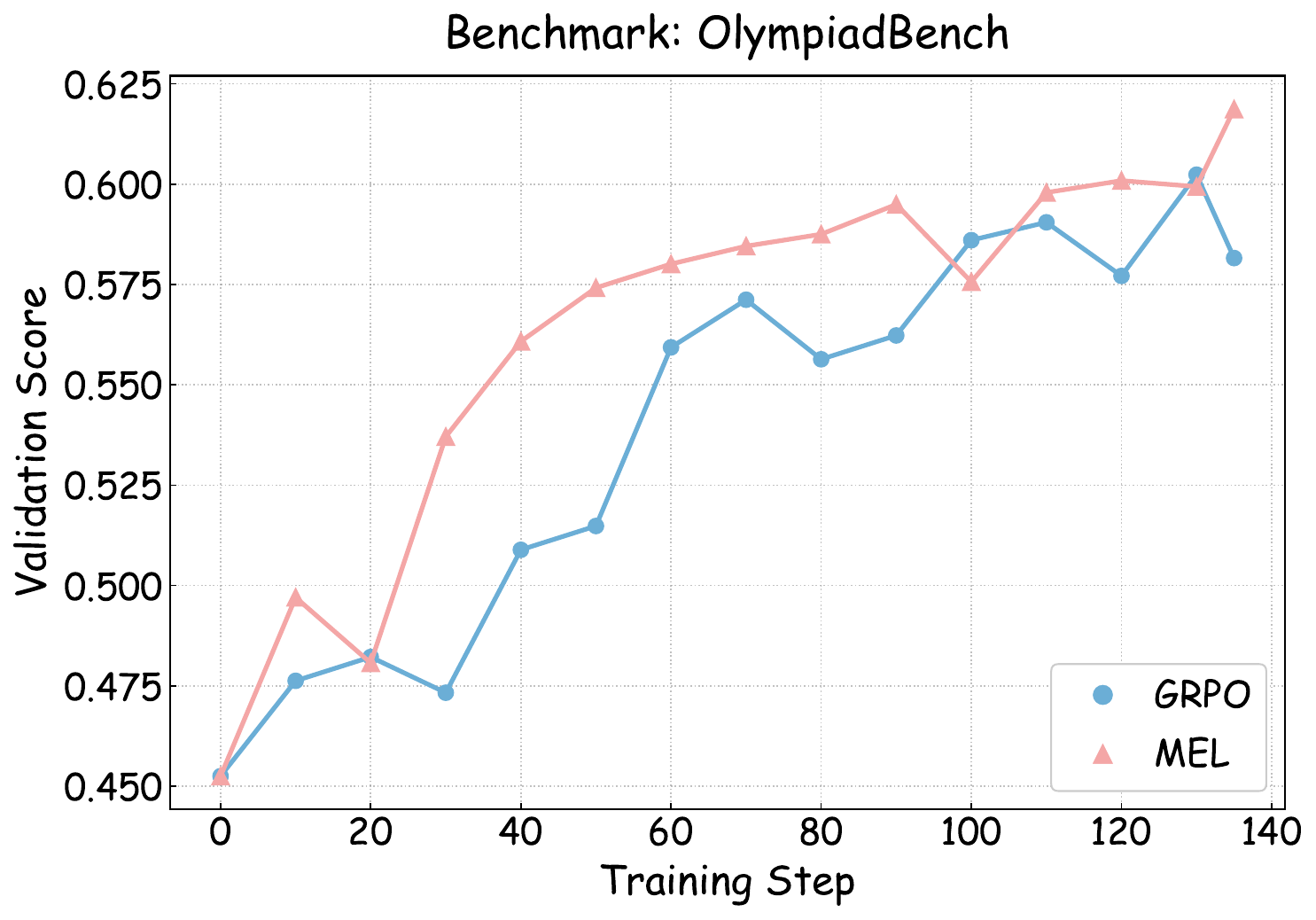}
  \end{minipage}
  \hfill
  \begin{minipage}{0.32\textwidth}
    \centering
    \includegraphics[width=\textwidth]{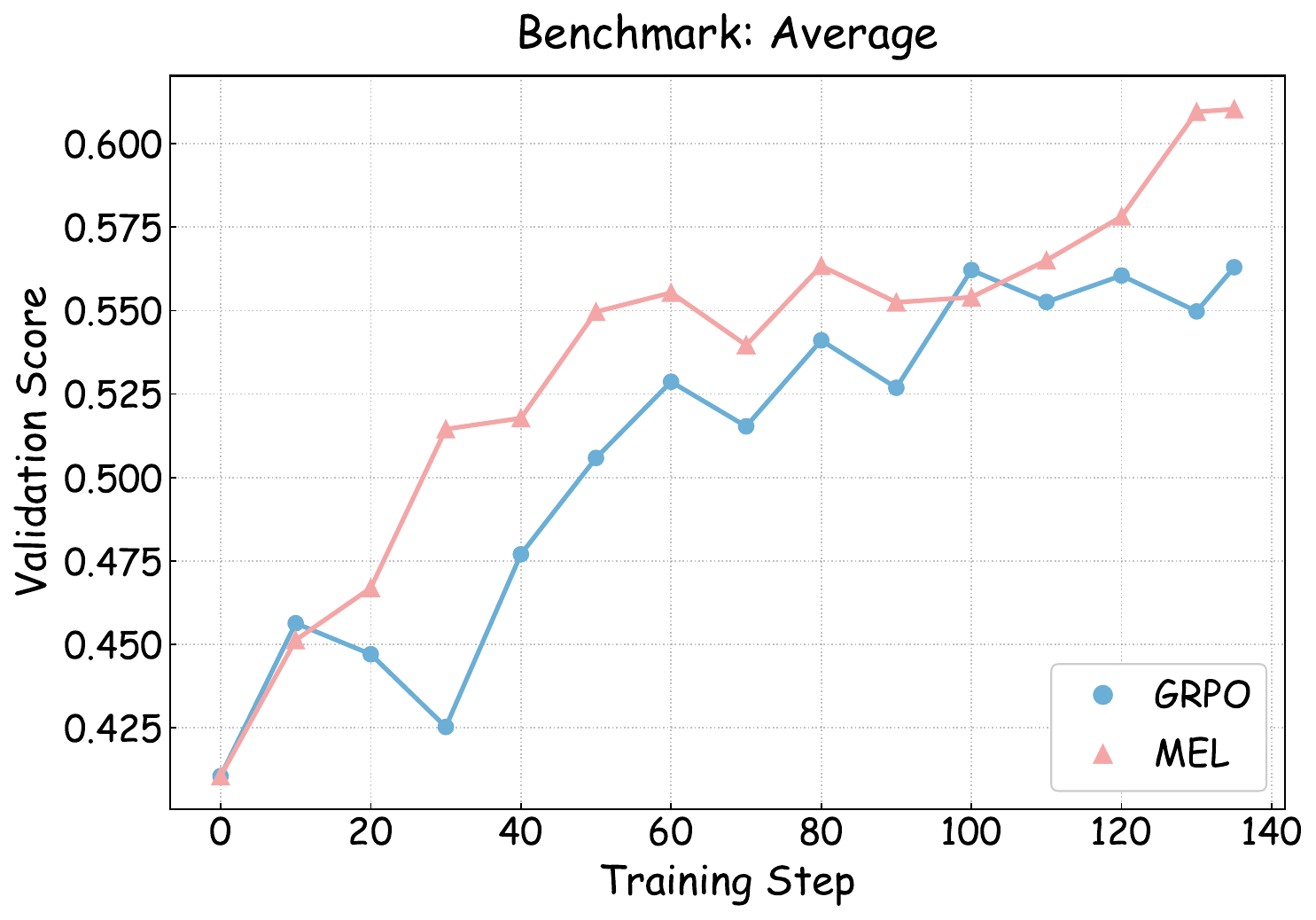}
  \end{minipage}

  \caption{Performance evolution of GRPO and \modelname on Qwen3-14B-Base across training steps on multiple benchmarks.}
  \label{fig:14b_performance_evolution}
\end{figure*}

As illustrated in Figures~\ref{fig:4b_performance_evolution}, \ref{fig:8b_performance_evolution}, and \ref{fig:14b_performance_evolution}, we visualize the performance evolution of models with different scales (Qwen3-4B-Base, Qwen3-8B-Base, and Qwen3-14B-Base) across multiple benchmarks throughout training.
It can be observed that \modelname consistently outperforms standard GRPO in terms of average performance on all benchmarks.

\vspace{100pt}
\section{Retention Ratio of Meta-Experience}
\begin{figure}[h]
\centering
\includegraphics[width=0.8\textwidth]{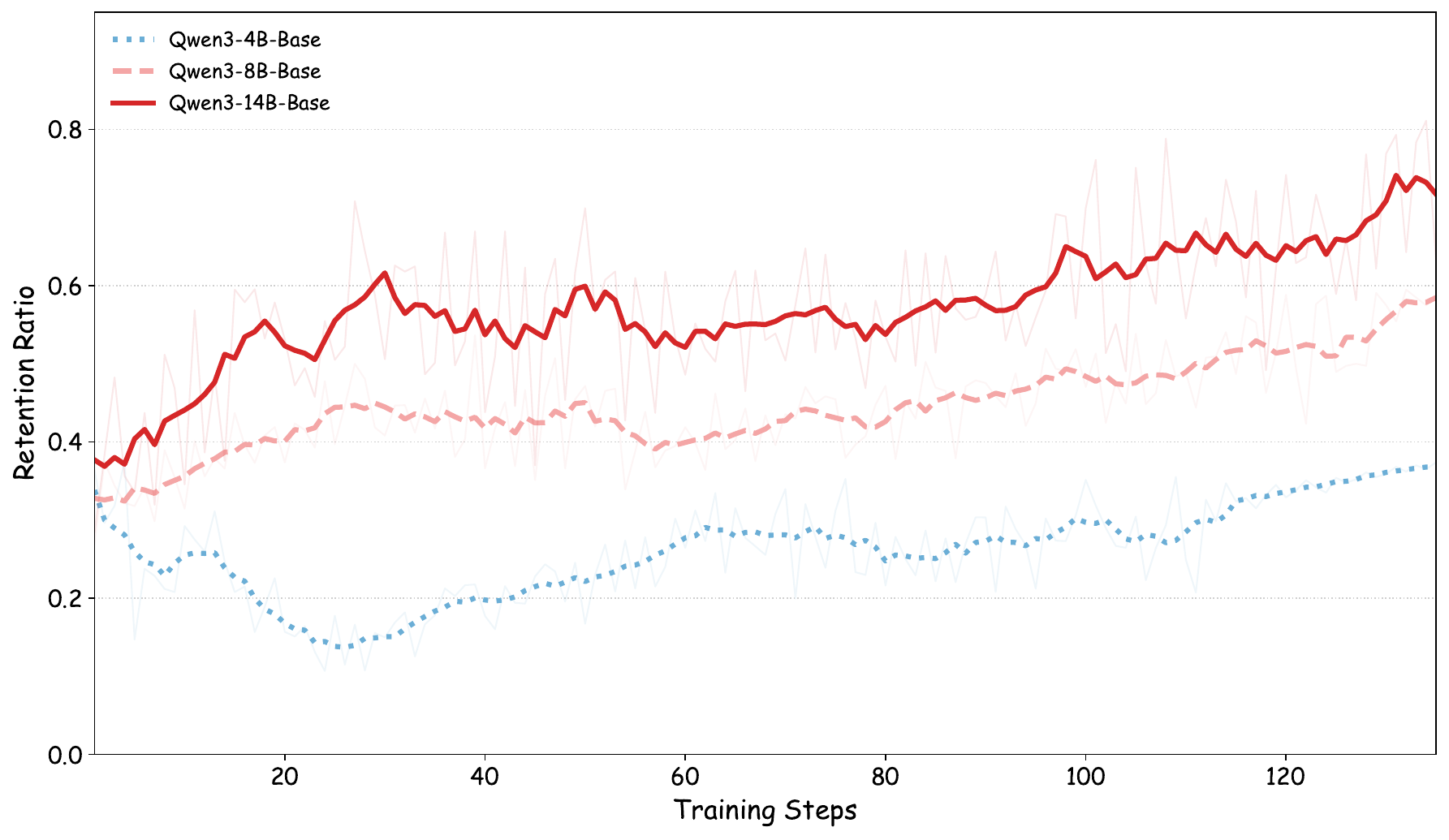}
  \caption{Dynamics of the retention ratio of \modelname across different model scales.}
  \label{fig:retention_ratio}
\end{figure}

Through empirical validation via replay, \modelname is able to collect high-quality meta-experiences.
To examine the utilization of collected meta-experiences, Figure~\ref{fig:retention_ratio} reports the retention ratio of meta-experiences after empirical validation throughout training.
We observe that the retention ratio consistently increases with model scale, indicating that larger models are more effective at abstracting high-quality knowledge into meta-experiences, thereby achieving higher retention.

\definecolor{mintblue}{RGB}{210,235,250}
\definecolor{mintframe}{RGB}{120,180,220} 
\definecolor{minttitle}{RGB}{100,150,200} 
\definecolor{minttext}{RGB}{50,80,120}    

\definecolor{runzhemilk}{RGB}{255,235,245} 
\definecolor{roseframe}{RGB}{230,120,150}  
\definecolor{runzhecotton}{RGB}{255,170,200}    
 

\newtcolorbox{promptbox}[1]{
enhanced,
breakable,
width=\textwidth,
colback=mintblue!40!white,
colframe=mintframe,
colbacktitle=minttitle!70!white,
coltitle=white,
title=\textbf{#1},
fonttitle=\bfseries,
sharp corners=south, 
borderline={0.8pt}{0pt}{minttitle},
boxrule=0.8pt,
arc=6pt, 
left=6pt, right=6pt, top=6pt, bottom=6pt,
before skip=10pt, after skip=10pt,
drop shadow=black!15, 
}
\clearpage
\section{Prompt Template}
We use the same prompt template for all models.
Details of the prompts used for meta-experience construction and for empirical validation via replay are shown below.

\begin{promptbox}{Meta-Experience Prompt}

You are a \textbf{Meta-Cognitive Reasoning Analyst} specializing in self-reflection, error root-cause analysis, and the extraction of generalizable heuristics.

You are provided with multiple solution trajectories for the same problem. Note that the labels \textbf{Correct} or \textbf{Incorrect} apply to the \textbf{final answer}, but the reasoning process itself may contain twists and turns.
Your task is to conduct a \textbf{deep comparative autopsy} of the thinking processes. You must identify the \textbf{structural} differences in cognition that led to success or failure, and synthesize these into abstract principles for future use.

\textbf{Core Analysis Requirements:}

\begin{enumerate}
  \item \textbf{Deep Dive into Correct Trajectories (Resilience \& Robustness Analysis):}
  \begin{itemize}
    \item \textbf{Scenario A (Self-Correction):} If you find the reasoning contains initial errors or uncertainties, look for moments of \textbf{self-correction}. What triggered the realization? What structural insight allowed the reasoning to pivot back to the right track?
    \item \textbf{Scenario B (Flawless Execution):} When every step of the reasoning is correct from the start, identify the \textbf{Foundational ``Immunity''}. What specific definition, clear knowledge representation, or disciplined step-by-step verification prevented this Agent from falling into the traps that the Incorrect Agent fell into?
    \item \textbf{Goal:} Extract the specific logic validation technique or robust mental representation that saved the solution.
  \end{itemize}

  \item \textbf{Deep Dive into Incorrect Trajectories (Vulnerability Analysis):}
  \begin{itemize}
    \item You must identify not only \textit{where} the math/logic went wrong, but also \textit{why} the reasoning drifted.
    \item Identify: The \textbf{``Bifurcation Point''} where a correct start turned into a hallucination or logic gap.
    \item Analyze: The \textbf{latent cognitive defect} (e.g., concept conflation, rigid mindset, overlooking edge cases, intuitive bias) that caused the error.
    \item Identify: What specific knowledge point or constraint was violated?
  \end{itemize}

  \item \textbf{Comparative Synthesis:}
  \begin{itemize}
    \item Contrast the \textbf{Solutions and Decision Boundaries}. Why did the successful trajectory avoid the trap that the failed one fell into?
    \item What \textbf{structural insight} did the winner have that the loser missed? (e.g., The winner treated the problem as a geometric issue, while the loser got stuck in algebra).
  \end{itemize}

  \item \textbf{Strict Generalization Constraint:}
  \begin{itemize}
    \item \textbf{Forbidden:} Do NOT mention the specific numbers, variables, or exact answer of the current problem in your ``Heuristics'' or ``Reflective Summary''.
    \item \textbf{Required:} Convert specific lessons into \textbf{abstract heuristics} (e.g., instead of ``The integral of $x^2$ is\ldots'', use ``When integrating polynomial functions\ldots''). Formulate them as conditionally triggered rules (``If\ldots Then\ldots'', ``When dealing with [Concept X]\ldots I should\ldots'').
  \end{itemize}
\end{enumerate}

\textbf{Output Format (Strict Adherence Required)}

\textbf{1. Failure Resolution Path \& Error Pattern Recognition} \textbf{(Mandatory for incorrect samples)}

\begin{itemize}
\item \textbf{Failure Point:} Identify the exact step where logic diverged. Did it start correctly? Where did the drift happen?
\item \textbf{Latent Cognitive Pattern:} Reveal the deep-seated reasoning defect. Was it a bias? A missing prerequisite? A misunderstanding of the prompt's intent? Do not list surface-level calculation errors.
\end{itemize}

\textbf{2. Analysis of Success Factors} \textbf{(Mandatory for correct samples)}
\begin{itemize}
\item \textbf{Reasoning Pivot (If applicable):} If the path involved self-correction, describe the moment of realization and the strategy used to fix it.
\item \textbf{Robustness Factor (If flawless):} If the path was perfect, explain the fundamental concept or structural approach that effectively ``immunized'' the reasoning against common errors.
\item \textbf{Reason for Effectiveness:} Why did this perspective work? What fundamental logic did it satisfy?
\end{itemize}

\textbf{3. First-Person Reflective Summary} \textbf{(Mandatory)}

Write a meta-cognitive reflection from the first-person perspective (``I'').

\begin{itemize}
\item \textbf{Review:} Briefly review the thinking process differences.
\item \textbf{Insight:} Discuss the specific knowledge point or cognitive habit that was critical.
\item \textbf{Action:} Explain how you will restructure your approach to avoid the identified ``Internal Reasoning Defects'' in the future.
\end{itemize}

Focus on the ``How'' of thinking, not the ``What'' of the answer.

\textbf{4. Subject Heuristics (Internalized Experience)} \textbf{(Mandatory)}

\begin{itemize}
\item \textbf{[Pattern Name]:} If [abstract condition] occurs, then [abstract action]\ldots
\item \textbf{[Pattern Name]:} When dealing with [concept type], I must strictly verify [constraint]\ldots
\end{itemize}
(Note: These must be applicable to *future* problems of a similar class, completely stripped of this problem's specifics.)

Here the question and the corresponding solutions.
$<$question$>$ \{question\} $<$/question$>$\\

Solution 1:\\
$<$answer$>$ \{error\_ans\} $<$/answer$>$\\
$<$judge$>$ Incorrect $<$/judge$>$\\

Solution 2:\\
$<$answer$>$ \{correct\_ans\} $<$/answer$>$\\
$<$judge$>$ Correct $<$/judge$>$

\end{promptbox}

\begin{promptbox}{Empirical Validation Prompt}
Prior study has provided some internal reference information relevant to this question, including the \textbf{key approaches, steps, and reasoning} needed for a correct solution; the \textbf{typical reasoning biases}, \textbf{logical flaws}, or \textbf{pitfalls} that appear in incorrect solutions; and various \textbf{heuristic insights} on how to complete this problem more effectively.  \\

\{experience\}\\

Now, please fully internalize this information as your own experience, then independently think through the problem in detail and produce a complete answer.

\textbf{Note:}
\begin{itemize}
  \item You must perform full, in-depth reasoning internally and arrive at the final answer while making full use of the information above.
\end{itemize}

\textbf{Answer the following question:}

\{question\}

\end{promptbox}

\end{document}